\documentclass[10pt,twocolumn,letterpaper]{article}

\usepackage{iccv}
\usepackage{times}
\usepackage{epsfig}
\usepackage{graphicx}
\usepackage{amsmath}
\usepackage{amssymb}
\usepackage{color}
\usepackage{multicol}
\usepackage{multirow}
\usepackage{makecell}
\usepackage{booktabs}
\usepackage{bbm}
\usepackage{bm} 
\usepackage{enumitem}
\usepackage{algorithm}
\usepackage{algorithmic}
\usepackage{amsthm}
\usepackage{caption}
\usepackage{subcaption}
\usepackage[numbers,sort]{natbib}
\usepackage{etoolbox}

\makeatletter
\pretocmd\@maketitle
  {\def\@makefnmark{\hbox{\@textsuperscript{\normalfont\@thefnmark}}}}
  {}{\FAIL}
\makeatother

\usepackage[pagebackref=true,breaklinks=true,letterpaper=true,colorlinks,bookmarks=false]{hyperref}
\usepackage{xcolor}
\newcommand{\myPara}[1]{\vspace{0.05in}\noindent\textbf{\hbox{#1~}}~\ignorespaces}

\definecolor{citecolor}{HTML}{0071BC}
\definecolor{linkcolor}{HTML}{ED1C24}
\definecolor{blue}{HTML}{0071BC}
\definecolor{red}{HTML}{ED1C24}

\hypersetup{colorlinks=true, linkcolor=linkcolor, citecolor=citecolor,urlcolor=black}
\iccvfinalcopy %

\def \xx {\bm{x}}
\def \zz {\bm{z}}
\DeclareMathOperator*{\argmax}{arg\,max}

\newcommand*\samethanks[1][\value{footnote}]{\footnotemark[#1]}

\begin{document}
\title{When Noisy Labels Meet Long Tail Dilemmas:\\ A Representation Calibration Method}

\author{%
\textbf{Manyi Zhang$^{1,}$\thanks{Equal contributions. This work was completed when the first two authors were interns guided by the last author.}
\quad 
Xuyang Zhao$^{2,}$\samethanks
\quad
Jun Yao$^{3}$
\quad Chun Yuan$^{1,}$\thanks{Correspondance to Weiran Huang (weiran.huang@outlook.com) and Chun Yuan (yuanc@sz.tsinghua.edu.cn).}
\quad
Weiran Huang$^{4,}$\samethanks}\\[0.3cm] 
$^{1}$SIGS, Tsinghua University \quad $^{2}$Peking University \quad 
$^{3}$Huawei Noah's Ark Lab \\[0.1cm] 
$^{4}$Qing Yuan Research Institute, SEIEE, Shanghai Jiao Tong University
}

\maketitle

\begin{abstract}
   Real-world large-scale datasets are both noisily labeled and class-imbalanced. The issues seriously hurt the generalization of trained models. It is hence significant to address the simultaneous incorrect labeling and class-imbalance, i.e., the problem of learning with noisy labels on long-tailed data. Previous works develop several methods for the problem. However, they always rely on strong assumptions that are invalid or hard to be checked in practice. In this paper, to handle the problem and address the limitations of prior works, we propose a representation calibration method RCAL. Specifically, RCAL works with the representations extracted by unsupervised contrastive learning. We assume that without incorrect labeling and class imbalance, the representations of instances in each class conform to a multivariate Gaussian distribution, which is much milder and easier to be checked. Based on the assumption, we recover underlying representation distributions from polluted ones resulting from mislabeled and class-imbalanced data. Additional data points are then sampled from the recovered distributions to help generalization. Moreover, during classifier training, representation learning takes advantage of representation robustness brought by contrastive learning, which further improves the classifier performance.
   We derive theoretical results to discuss the effectiveness of our representation calibration. Experiments on multiple benchmarks justify our claims and confirm the superiority of the proposed method. 
   
\end{abstract}

\section{Introduction}
\label{sec:intro}
\begin{figure}[!t]
    \centering
    \includegraphics[width=0.95\linewidth]{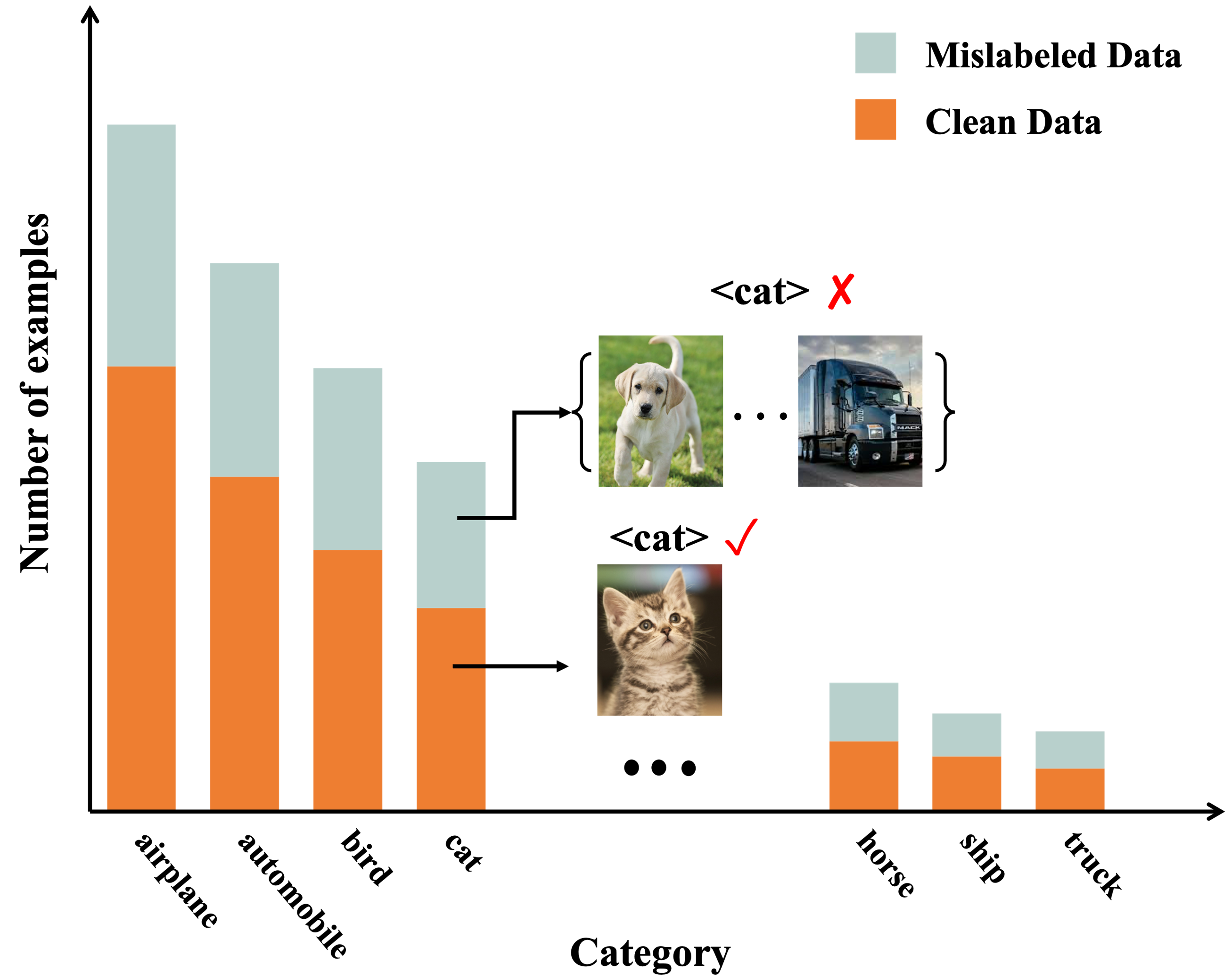} 
    \caption{The illustration of the problem setup. The observed data exhibit a long-tailed distribution. The number of clean data and mislabeled data varies in each class. }
    \label{fig:problem_setting}
    \vspace{-12pt}
\end{figure}

Deep learning has made rapid progress in many fields~\cite{goodfellow2016deep}, primarily driven by large-scale and high-quality annotated datasets~\cite{lecun2015deep,cao2020heteroskedastic,han2018co,yao2019searching,wu2021class2simi,li2022estimating}. Unfortunately, it is hard to obtain such perfect datasets in practice, mainly from two aspects: (1) a part of data is wrongly labeled due to its intrinsic ambiguity and mistakes of annotators~\cite{ma2020normalized,li2022selective,xia2019anchor,wei2021optimizing,liang2022few}; (2) data is class-imbalanced, where a long-tailed class distribution exhibits~\cite{wang2021label,zhong2021improving,hu2020learning}. In real-world settings, both imperfect situations usually coexist (see Figure~\ref{fig:problem_setting}). For example, the WebVision dataset~\cite{li2017webvision}, a large-scale image dataset crawled from the web, contains about $20\%$ mislabeled data. Meanwhile, the number of examples in the most frequent class is over 20 times that of examples in the most scarce class~\cite{karthik2021learning}.

Although many previous works have emerged to address the problems of learning with noisy labels and learning with long-tailed data separately, they cannot work well when the two imperfect situations exist simultaneously. Namely, they are weak for learning with noisy labels on long-tailed data. Concretely, the methods specialized for learning with noisy labels always rely on some assumptions. Nevertheless, the assumptions are invalid due to the long-tailed issue. For example, the popularly used memorization effect~\cite{han2018co} for tackling noisy labels cannot be applied, since clean data belonging to tail classes show similar training dynamics to those mislabeled data, \textit{e.g.}, similar training losses~\cite{cao2020heteroskedastic,xia2021sample}. Also, the noise transition matrix used for handling noisy labels cannot be estimated accurately. This results from that the relied anchor points of tail classes cannot be identified from noisy data, as the estimations of noisy class posterior probabilities for tail classes are not accurate. Moreover, the methods specialized for learning with long-tailed data mainly adopt re-sampling and
re-weighting techniques to balance the classifier. The side-effect of mislabeled data is not taken into consideration, which results in the accumulation of label errors.

The weaknesses of the above specialized methods motivate us to develop more advanced methods for the realistic problem of learning with noisy labels on long-tailed data. Existing methods targeting this problem can be divided into two
main categories. The methods in the first category are to distinguish mislabeled data from the data of tail classes for follow-up procedures. However, the distinguishment is adversely affected by mislabeled data, since the information used for the distinguishment comes from deep networks that are trained on noisy long-tailed data. The methods in the second category are to reduce the side-effects of mislabeled data and long-tailed data in a unified way, relying on strong assumptions. For example, partial data should have the same aleatoric uncertainty~\cite{cao2020heteroskedastic}, which is hard to check in practice.

In this paper, we focus on this realistic problem: learning with noisy labels on long-tailed data. To address the issues of prior works, we propose a \textit{representation calibration} method named RCAL. Generally, RCAL works on the level of deep representations, \textit{i.e.}, extracted features by deep networks for instances. Technically, we first employ unsupervised contrastive learning to achieve representations for all training instances. As the procedure of representation learning is not influenced by corrupted training labels, the achieved representations are naturally robust~\cite{zheltonozhskii2022contrast, xue2022investigating, ghosh2021contrastive}. Afterward, based upon the achieved representations, two representation calibration strategies are performed: distributional and individual representation calibrations. 

In more detail, the distributional representation calibration aims to recover representation distributions before data corruption. Specifically, we assume that before training data are corrupted, the deep representations of instances in each class conform to a multivariate Gaussian distribution. Compared to the previously mentioned assumptions, the assumption used in this paper is much milder. Its rationality is also justified by many works~\cite{yang2021free, wang2021label, zhang2022tackling}. With a density-based outlier detector, robust estimations of multivariate Gaussian distributions are obtained. Moreover, since the insufficient data of tail classes may cause biased distribution estimations, the statistics of distributions from head classes are employed to calibrate the estimations for tail classes. After the distributional calibration for all classes, we sample multiple
data points from the recovered distributions, which makes training data more balanced
and helps generalization\footnote{Perhaps in some actual scenes, the deep representations cannot conform to multivariate Gaussian distributions perfectly. We show that, based on the assumption of multivariate Gaussian distributions, it is enough to get state-of-the-art classification performance using sampled 
data points from estimated multivariate Gaussian distributions. The empirical evidence on real-world datasets is provided in Section~\ref{sec:exp}.}. As for individual representation calibration, considering that the representations obtained by contrastive learning are robust, we restrict that the subsequent learned representations during training are close to them. The individual representation calibration implicitly reduces the hypothesis space of deep networks, which mitigates their overfitting of mislabeled and long-tailed data. Through the above procedure of representation calibration, the learned representations on noisy long-tailed data are calibrated towards uncontaminated representations. The robustness of deep networks is thereby enhanced with such calibrated representations, following better classification performance. 

The contributions of this paper are listed as follows: (1) We focus on learning with noisy labels on long-tailed data, which is a realistic but challenging problem. The weaknesses of previous works are carefully discussed. (2) We propose an advanced method RCAL for learning with noisy labels on long-tailed data. Our method benefits from the representations by contrastive learning, where two types of representation calibration strategies are proposed to improve network robustness. (3) We derive theoretical results to confirm the effectiveness of our calibration strategies under some conditions. (4) We conduct extensive experiments on both simulated and real-world datasets. The results demonstrate our representation calibration method's superiority over existing state-of-the-art methods. In addition, detailed ablation studies and discussions are provided.

\section{Related Works}
\label{sec:related}
\myPara{Learning with noisy labels.}
There is a series of works proposed to deal with noisy labels, which includes but is not limited to estimating the noise transition matrix~\cite{xia2020part,chengclass,yao2020dual}, selecting confident examples~\cite{peng2020large,wang2022scalable,patel2023adaptive}, reweighting examples~\cite{ren2018learning,liu2015classification}, and correcting wrong labels~\cite{liu2022adaptive}. Additionally, some state-of-the-art methods combine multiple techniques, \textit{e.g.},  DivideMix~\cite{li2020dividemix}, ELR+~\cite{liu2020early}, and Sel-CL+~\cite{li2022selective}.

\myPara{Learning with long-tailed data.}
Existing methods tackling long-tailed data mainly focus on: (1) re-balancing data distributions, such as over-sampling~\cite{haixiang2017learning,byrd2019effect,peng2020large}, under-sampling~\cite{drummond2003c4,byrd2019effect,he2009learning}, and class-balanced sampling~\cite{mahajan2018exploring,shen2016relay}; (2) re-designing loss functions, which includes class-level re-weighting~\cite{cui2019class,cao2019learning,hong2021disentangling,li2022equalized,wang2021adaptive,wang2021seesaw} and instance-level re-weighting~\cite{lin2017focal,shu2019meta,ren2018learning,zhao2022adaptive}; (3) decoupling representation learning and classifier learning~\cite{kang2019decoupling,zhang2021distribution}; (4) transfer learning from head knowledge to tail classes~\cite{hu2020learning,liu2021gistnet,he2021distilling}.

\myPara{Learning with noisy labels on long-tailed data.}
A line of research has made progress towards simultaneously learning with imbalanced data and noisy labels. CurveNet~\cite{jiang2022delving} exploits the informative loss curve to identify different biased data types and produces proper example weights in a meta-learning manner, where a small additional unbiased data set is required. HAR~\cite{cao2020heteroskedastic} proposes a heteroskedastic adaptive regularization approach to handle the joint problem in a unified way. The examples with high uncertainty and low density will be assigned larger regularization strengths. RoLT~\cite{wei2021robust} claims the failure of the small-loss trick in long-tailed learning and designs a prototypical error detection method to better differentiate the mislabeled examples from rare examples. TBSS~\cite{zhang2022combating} designs two metrics to detect mislabeled examples under long-tailed data distribution. A semi-supervised technique is then applied.

\section{Methodology}
\label{sec:method}

\subsection{Preliminaries}

\myPara{Notation.} In the sequel, scalars are in lowercase letters. Vectors are in lowercase boldface letters. Let $[z]=\{1,2,\ldots,z\}$. Besides, $|\mathcal{B}|$ denotes the total number of elements in the set $\mathcal{B}$.

\myPara{Problem setup.} We consider a $K$-class classification problem, where $K\geq 2$. We are given an imbalanced and noisily labeled training dataset $\tilde{\mathcal{S}}=\{(\xx_i, \tilde{y}_i)\}_{i=1}^{n}$, where $n$ is the sample size, $\xx_i$ denotes the $i$-th instance and its label $\tilde{y}_i\in [K]$ may be incorrect. For the label $\tilde{y}_i$, the corresponding true label is denoted by $y_i$, which is unobservable. Let the number of training data belonging to $k$-th class be $n_k$.  Without loss of generality, we suppose that the classes are sorted in decreasing order, based on the number of training data in each class, \textit{i.e.}, $n_1\geq...\geq n_{K}$. Afterward, all classes can be recognized into two parts: head classes (referred as $\mathcal{G}_h$) and tail classes (referred as $\mathcal{G}_t$). In this paper, the aim is to learn a classifier robustly by \textit{only using the imbalanced and noisily labeled training dataset}, which can infer proper labels for unseen instances. 

\begin{figure}[!t]
    \centering
    \includegraphics[scale=0.39]{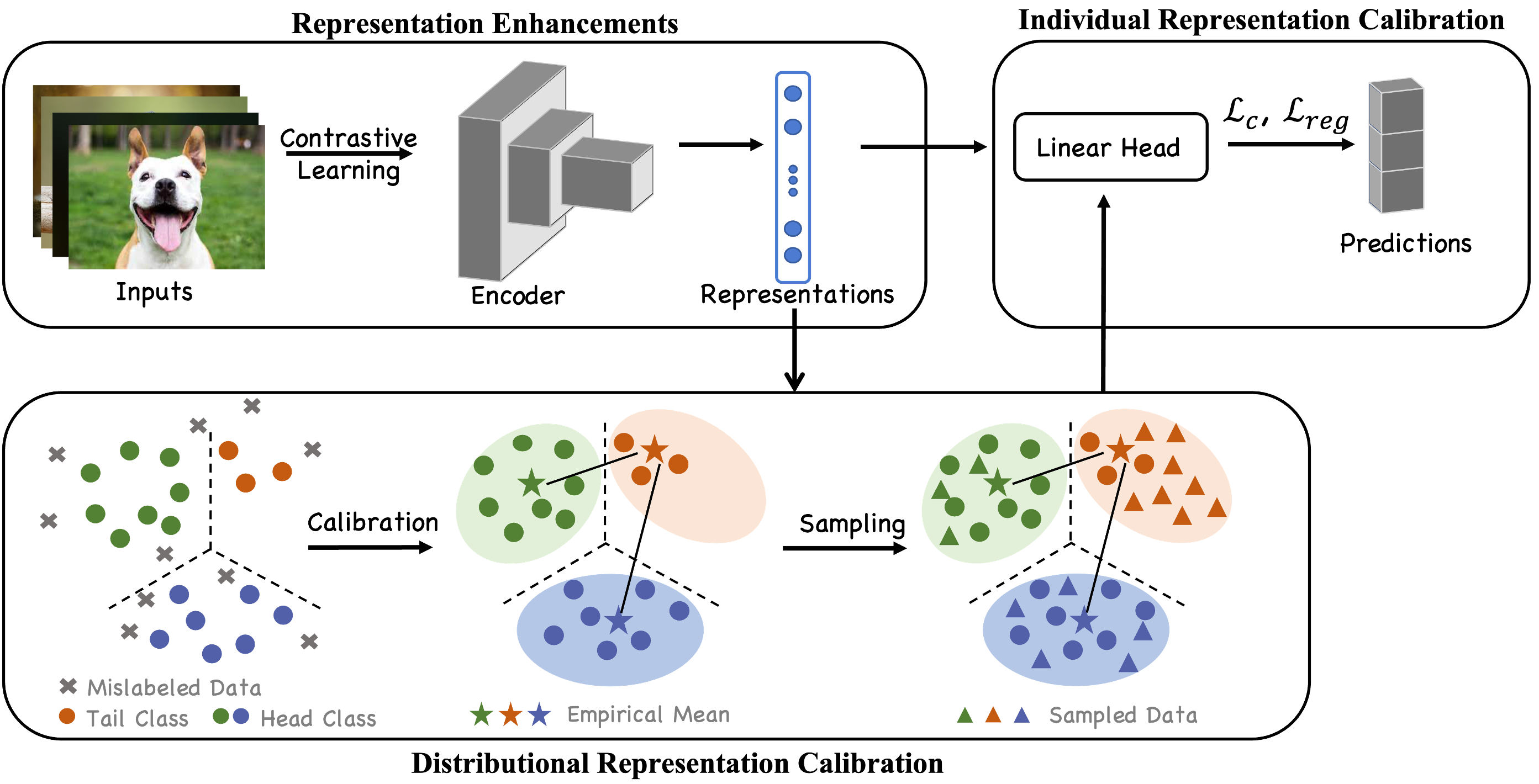} 
    \caption{The illustration of the proposed method, uses the representations achieved by contrastive learning for follow-up distributional and individual representation calibrations.}
    \label{fig:overview}
    \vspace{-12pt}
\end{figure}

\myPara{Algorithm overview.} In the following, we discuss the proposed method RCAL step by step. Generally, RCAL consists of two stages: (1) the stage of representation enhancements by contrastive learning; (2) the stage of improving the classifier's robustness by representation calibration, which is performed with before enhanced representations. The procedure of RCAL is illustrated in Figure~\ref{fig:overview}. We provide more technical details of our method as follows. 

\subsection{Enhancing Representations Through Contrastive Learning}\label{sec:3.2}
To improve the robustness of deep representations of instances for handling noisy labels in long-tailed cases, we exploit self-supervised contrastive learning. Intuitively, as the representation learning in self-supervised contrastive learning does not access the labels of training data, the achieved representations will not be influenced by incorrect labels~\cite{xue2022investigating}. Besides, prior work~\cite{liu2021self} shows that contrastive learning can improve the network tolerance to long-tailed data. Therefore, deep representations achieved with contrastive learning help tackle noisy labels in long-tailed cases.

Specifically, we utilize the encoder networks following the popular setup in MOCO~\cite{chen2020improved}. For an input $\xx$, we apply two random augmentations and thus generate two views $\xx^q$ and $\xx^k$. The two views are then fed into a query encoder $f(\cdot)$ and a key encoder $f^{\prime}(\cdot)$, which  generates representations $\zz^q = f(\xx^q)$ and $\zz^k=f^{\prime}(\xx^k)$. Thereafter, a projection head, \textit{i.e.}, a 2-layer MLP, maps the two representations to lower-dimensional embeddings $\hat{\zz}^q$ and $\hat{\zz}^k$. MOCO also maintains a large queue to learn good representations. The key encoder uses a momentum update with the query encoder to keep the queue as consistent as possible. The contrastive loss for the input $\xx_i$ can be expressed as:
\begin{equation}\label{eq:contrastive}
    \mathcal{L}_{con}(\xx_i)= -\log \frac{\exp(\hat{\zz}_i^q\cdot\hat{\zz}_i^{k}/\tau)}{\Sigma_{\hat{\zz}^{k^\prime}\in \mathcal{A}}\exp(\hat{\zz}_i^q\cdot\hat{\zz}^{k^{\prime}}/\tau)},
\end{equation}
where $\mathcal{A}$ is the queue, and  $\tau>0$ is a temperature parameter. The enhanced representation of the input $\xx_i$ is achieved by minimizing the loss in Eq.~(\ref{eq:contrastive}). The representation $\zz^q=f(\xx)$ (simplified as $\zz$) is extracted from the query encoder for later representation distribution calibration.

\subsection{Distributional Representation Calibration}
In this paper, we assume that before corruption by class-imbalanced noisy labels, the deep representations of training data in each class conform to a multivariate Gaussian distribution. Note that the assumption is applied to deep representations but not original instances, since deep representations are more informative for following procedures. Besides, the assumption is mild and has been verified by existing works~\cite{yang2021free,xian2018feature,zhang2022tackling,wang2021label}.

For $K$ classes, we have $K$ multivariate Gaussian distributions. The distribution belonging to the $k$-th class is denoted by $\mathcal{N}(f(\xx) | \bm{\mu}_k, \bm{\Sigma}_k)$ with $\bm{\mu}_k\in\mathbbm{R}^m$ and $\bm{\Sigma}_k\in\mathbbm{R}^{m\times m}$, where $m$ denotes the dimension of the deep representation. Although without introducing class labels in representation learning by contrastive learning, there is a clustering effect for the obtained representations~\cite{zheltonozhskii2022contrast}. Therefore, we can exploit them for modeling the multivariate Gaussian distributions at a class level. Due to the side-effect of mislabeled data, the prior multivariate Gaussian distributions are corrupted. As mentioned, we tend to tackle noisy labels in long-tailed cases, with representation distribution calibration. Therefore, we need to estimate the multivariate Gaussian distributions that are not affected by mislabeled data. 

\myPara{Robust estimations of Gaussian distributions.} If deep representations are not contaminated due to mislabeled data, the empirical mean has a $L_2$-error at most $\mathcal{O}(\sqrt{m/n_k})$ from the true mean $\bm{\mu}_k$. Owing to the existence of noisy labels, the empirical estimation fails. We therefore develop an advanced estimation method. Note that the representations learned from contrastive learning are clustered among similar representations and not influenced by noisy labels. They hence can help detect outliers in the representation space for the estimations of Gaussian distributions.

Technically, given the learned representations $\zz$, we employ the Local Outlier Factor (LOF) algorithm~\cite{breunig2000lof} to detect outliers. The outliers are then removed for the following estimation. After performing the LOF algorithm on $\{(\zz_i, \Tilde{y}_i)\}_{i=1}^n$, we segregate clean data for each class. The set of preserved examples for the $k$-th class is denoted by $\tilde{\mathcal{S}}'_k$, where $\tilde{\mathcal{S}}'_k = \{(\zz_i, \Tilde{y}_i)\}_{i=1}^{|\tilde{\mathcal{S}}'_k|}$ with $|\tilde{\mathcal{S}}'_k|<n_k$. With $\tilde{\mathcal{S}}'_k$, we estimate the multivariate Gaussian distribution as 
\begin{align*}
    &\hat{\bm{\mu}}_k = \sum_{\{i|(\zz_i,\tilde{y}_i)\in\tilde{\mathcal{S}}'_k\}}\frac{\zz_i}{|\tilde{\mathcal{S}}'_k|},\\
    &\hat{\bm{\Sigma}}_k = \sum_{\{i|(\zz_i,\tilde{y}_i)\in\tilde{\mathcal{S}}'_k\}}\frac{(\zz_i-\hat{\bm{\mu}}_k)(\zz_i-\hat{\bm{\mu}}_k)^\top}{|\tilde{\mathcal{S}}'_k|-1},
\end{align*}
where the mean of representation vectors is calculated as the mean of every single dimension in the vector. 

\myPara{Further calibration for tail classes.} As the size of the training data belonging to tail classes is small, it may not be enough for accurately estimating their multivariate Gaussian distributions with the above robust estimation. Inspired by similar classes having similar means and covariance on representations~\cite{yang2021free, wang2021label}, we further borrow the statistics of head classes to assist the calibration of tail classes. Specifically, we measure the similarity by computing the Euclidean distances between the means of representations of different classes. For the tail class $k$, we select top $q$ head classes with the closest Euclidean distance to the mean $\hat{\bm{\mu}}_k$:
\begin{align*}
    \mathcal{B}_k &= \left\{-||\hat{\bm{\mu}}_i-\hat{\bm{\mu}}_k||^2\ \big|\  i\in\mathcal{G}_h\right\},\\
    \mathcal{C}^q_k &=\left\{i\ \big| -||\hat{\bm{\mu}}_i-\hat{\bm{\mu}}_k||^2 \in \texttt{topq}(\mathcal{B}_k) \right\}. 
\end{align*}

Afterward, we can rectify the means and covariances of tail classes as follows:
\begin{align*}
    \omega_c^k & = \frac{n_c||\hat{\bm{\mu}}_c-\hat{\bm{\mu}}_k||^2}{\sum_{j\in \mathcal{C}^q_k}n_j||\hat{\bm{\mu}}_j-\hat{\bm{\mu}}_k||^2},\\
    \hat{\bm{\mu}}_k^{\prime} &= \gamma\sum_{c\in \mathcal{C}^q_k} \omega_c^k\hat{\bm{\mu}}_c + (1-\gamma)\hat{\bm{\mu}}_k,\\
     \hat{\bm{\Sigma}}_k^{\prime} & = \gamma\sum_{c\in \mathcal{C}^q_k}\omega_c^k\hat{\bm{\Sigma}}_c+(1-\gamma)\hat{\bm{\Sigma}}_k + \alpha \mathbf{1},
\end{align*}
where $\omega_c^k$ is the weight that is about using the statistics of the head class $c$ to help the calibration of the tail class $k$. The head classes that are more similar to the tail class $k$ will be endowed with smaller weights. Additionally, $\gamma$ is the confidence on the statistics computed from head classes, $\mathbf{1}\in\mathbbm{R}^{m\times m}$ is the matrix of ones, and $\alpha \in \mathbbm{R}^{+}$ is a hyperparameter that controls the degree of disturbance. The application of the disturbance can make the estimations of covariances more robust. At last, the multivariate Gaussian distributions for head classes are achieved by $\mathcal{N}(\zz|\hat{\bm{\mu}},\hat{\bm{\Sigma}})$, while the multivariate Gaussian distributions for tail classes are calibrated and given by $\mathcal{N}(\zz|\hat{\bm{\mu}}^{\prime}, \hat{\bm{\Sigma}}^{\prime})$.

After recovering all multivariate Gaussian distributions, we sample multiple data points from them for classifier training. As the recovered distributions are close to the representation distributions of clean data, training with these sampled data points can make the classifier more robust. Furthermore, we can control the number of sampled data points from different classes to make training data more balanced, which helps generalization.

\subsection{Individual Representation Calibration}
Before this, we finish distributional representation calibration to recover the multivariate Gaussian distributions. Going a further step to improve the representation robustness, we perform individual representation calibration which includes two parts.

First, considering that self-supervised contrastive learning provides us robust representations (Section~\ref{sec:3.2}), we restrict the distance between subsequent learned representations and the representations brought by contrastive learning. Specifically, we denote the representations brought by contrastive learning as $\zz^0$. Then the distance restriction is formulated as 
\begin{equation*}
    \mathcal{L}_{reg}(\xx) = ||\zz-\zz^0||^2 = ||f(\xx)-\zz^0||^2. 
\end{equation*}

Second, to further make learned representations robust to tackle noisy labels in long-tailed cases, we employ the \textit{mixup} method~\cite{zhang2017mixup}. Let the cross-entropy loss for the example $(\xx,\tilde{y})$ be $\mathcal{L}_c(\xx,\tilde{y})$. In the procedure of mixup, each time we randomly sample two examples $(\xx_i,\tilde{y}_i)$ and $(\xx_j,\tilde{y}_j)$, weighted combinations of these
two examples are generated as 
\begin{equation*}
    \xx_{i,j} = \lambda \xx_i + (1-\lambda)\xx_j\ \text{and}\ \tilde{y}_{i,j} = \lambda \tilde{y}_i + (1-\lambda)\tilde{y}_j,
\end{equation*}
where $\lambda\in [0,1]$ is drawn from the Beta distribution. Accordingly, the original training objective based on the cross-entropy loss is replaced with $\mathcal{L}_c(\xx_{i,j},\tilde{y}_{i,j})$. Note that, for the classification, we add a linear head $h$. The classification results of the instance $\xx$ are $h(f(\xx))$. Finally, the overall objective is formulated as 
\begin{equation}\label{eq:total_loss}
    \mathcal{L} = \mathcal{L}_c + \beta \mathcal{L}_{reg},
\end{equation}
where $\beta$ controls the strength of distance regularization. The algorithm flow of our method is provided in Algorithm~\ref{alg}. 
 
\begin{algorithm}[t]
    \algsetup{linenosize=\small} \small
    \caption{Algorithm of the proposed method RCAL}
    \label{alg}
    \begin{algorithmic}[1]
        \REQUIRE the training dataset $\Tilde{\mathcal{S}}=\{(\xx_i, \Tilde{y}_i\})_{i=1}^n$, regularization strength $\beta$, scalar temperature $\tau$, confidence weight $\gamma$, the pre-training epochs $T_{p}$, max epochs $T_{m}$.
        \FOR{$t=1,...,T_{p}$}
            \STATE \textbf{Pre-train} the encoder network $f$ with      MoCo~\cite{he2020momentum}.            
        \ENDFOR
        \STATE \textbf{Extract} deep representations of instances with $\zz=f(\xx)$.
        \FOR{$c=1,...,K$}
        \STATE \textbf{Perform} the LOF algorithm for the $c$-th class and obtain preserved examples $\tilde{\mathcal{S}}'_c$. 
        \STATE \textbf{Build} the multivariate Gaussian distribution $\mathcal{N}(f(\xx)|\hat{\bm{\mu}}_c,\hat{\bm{\Sigma}}_c)$ for $c$-th class using $\tilde{\mathcal{S}}'_c$. 
        \ENDFOR
        \STATE \textbf{Calibrate} the multivariate Gaussian distributions of tail classes with the statistics of head classes. 
        \STATE \textbf{Sample} data points from achieved multivariate Gaussian distributions of all classes.
        \FOR{$t=T_{p}+1, ..., T_{m}$}
            \STATE \textbf{Add} distance constraints between learned representations and representations brought by contrastive learning. 
            \STATE \textbf{Adopt} the mixup technology to original examples. 
            \STATE\textbf{Train} the encoder $f$ and the linear head $h$ simultaneously on the training dataset and sample data points with the training loss in Eq.~(\ref{eq:total_loss}).
        \ENDFOR
        \RETURN The robust classifier $h(f(\xx))$ for testing.
    \end{algorithmic}
\end{algorithm}

\newtheorem{definition}{Definition}[section]
\newtheorem{assumption}{Assumption}[section]
\newtheorem{example}{Example}[section]
\newtheorem{theorem}{Theorem}[section]
\newtheorem{corollary}{Corollary}[theorem]
\newtheorem{lemma}[theorem]{Lemma}
\newtheorem{proposition}[theorem]{Proposition}
\newtheorem{remark}{Remark}

\subsection{Theoretical Analysis}
We give a theoretical analysis to show the benefit of calibration. 
We begin by formally presenting some model assumptions of the tuples $\{(\bm{z}_i,\tilde{y}_i,y_i)\}_{i=1}^n$, where $\bm{z}_i$ is the deep representation, $y_i$ is the true label, and $\tilde{y}_i$ is the contaminated label. Note that $y_i$ is unobserved. For theoretical simplicity, we assume that $n_k=n_{\text{tail}}$ for each $k\in\mathcal{G}_t$ and $n_k=n_{\text{head}}$ for each $k\in\mathcal{G}_h$.

\begin{assumption}\label{assumption}
(1) (Gaussian deep representations) 
The $m$-dimensional representation $\bm{z}$ and the corresponding true label $y$ satisfies $P(y=k)=n_k/n$ and $\bm{z}\,|\,y = k \sim \mathcal{N}(\bm{\mu}_k, \bm{\Sigma})$. 

\vspace{0.1cm}
\noindent (2) (Class imbalance) There is a constant $\rho > 1$ such that $n_{\text{head}} \geq \rho \cdot n_{\text{tail}}.$

\vspace{0.1cm}
\noindent (3) (Random label flipping) There is a constant $\eta>0$ such that given the true label $y_i=k$, the contaminated label $\tilde{y}_i$ satisfies $P(\tilde{y}_i=j\,|\,y=k) = \eta\cdot n_j / n$ for $j\not=k$.

\vspace{0.1cm}
\noindent (4) (Informative head classes) There is a constant $\delta_q$ (depending on $q$) such that $\max_{j\in\mathcal{C}_k^q}\|\bm{\mu}_j - \bm{\mu}_k\| \leq \delta_q.$
\end{assumption}

For simplicity, in Assumption \ref{assumption} (1), we assume that all classes have the same covariance matrix $\bm{\Sigma}$. 
The $\rho$ introduced in Assumption \ref{assumption} (2) is the \textit{class imbalance ratio.}
A larger $\rho$ implies a more unbalanced sample size distribution.
The $\eta$ in Assumption \ref{assumption} (3) is the \textit{noise rate}, which measures the degree of label noise. 
Note that the label flipping probability is assumed to be proportional to the sample size. 
This comes from the intuition that people are more likely to misclassify labels into classes with larger sample size.
Assumption \ref{assumption} (4) is imposed to measure the extent to which head classes can help the estimation of tail classes.

For the Gaussian model \ref{assumption} (1), the Bayes optimal classifier on top of $\bm{z}$ is well-known to be Fisher's linear discriminant \cite{anderson2003multivariate}, which is defined as 
$$
h_*(\bm{z}) := \argmax_{k\in[K]} \big\{\log(n_k/n) + {\bm{\mu}}_k^\top {\bm{\Sigma}}^{-1}{\bm{z}} - \frac{1}{2}{\bm{\mu}}_k^\top \bm{\Sigma}^{-1} \bm{\mu}_k \big\}.
$$
In practice, the true mean $\bm{\mu}$ is unknown and needs to be estimated from data, whose estimation error directly affects the corresponding classification error. 
Therefore, to study the benefit of calibration, we give the estimation error of the calibrated mean $\hat{\bm{\mu}}'$ and the vanilla empirical mean $\hat{\bm{\mu}}$ in the following theorem.
\begin{theorem}
Under Assumption \ref{assumption}, there exists constant $C$ such that
\begin{equation}\label{excess-risk}
\mathbb{E}\|\hat{\bm{\mu}}_k - \bm{\mu}_k\|^2\leq C\cdot\left[ \eta^2 + \frac{m}{n_{\text{tail}}}\right]
\end{equation}
and
\begin{equation}\label{excess-risk-cali}
\mathbb{E}\|\hat{\bm{\mu}}'_k - \bm{\mu}_k\|^2\! 
\leq\! C\cdot\left[ \eta^2 + \delta_q^2 + \max\{\frac{m}{q\gamma{n}_{\text{tail}}}, \frac{m}{{n}_{\text{head}}}\}\right].\!
\end{equation}
\end{theorem}

The term $\eta^2$ appears in both \eqref{excess-risk} and \eqref{excess-risk-cali}, which is caused by the label noise in the training data and is inevitable.
The second term $\delta^2_q$ in \eqref{excess-risk-cali} is the bias introduced by using head classes to calibrate tail classes since they have different means. It is not involved in the vanilla classifier $\hat{h}$.
The last terms in \eqref{excess-risk} and \eqref{excess-risk-cali} are variance terms caused by the finite sample issue.
For the vanilla classifier, since the sample sizes of tail classes are relatively small, the variance term is dominated by $m/n_{\mathrm{tail}}$ and could be very large.
For the calibrated classifier, we can see that this term is significantly reduced if $\gamma$ or $q$ is large, since we can borrow strength from head classes whose sample sizes are large.

We would like to highlight the fact that, based on the deep representations pretrained by linear classifiers are sufficient to obtain good downstream performance. Thus, we consider the linear case over deep representations in this work, which is also adopted by most related theory papers. Moreover, our linear theory still provides insights into how calibration helps with long-tailed noisy tasks, making our method more reliable than other heuristic methods.
\section{Experiments}
\label{sec:exp}

\subsection{Baselines}\label{sec:4.1}
For comprehensive evaluations, we employ three types of comparison methods as follows: (1) Methods designed for learning with long-tailed data include LDAM~\cite{cao2019learning}, LDAM-DRW~\cite{cao2019learning}, CRT~\cite{kang2019decoupling}, NCM~\cite{kang2019decoupling} and MiSLAS~\cite{zhong2021improving}; (2) Methods designed for learning with noisy labels include Co-teaching~\cite{han2018co}, CDR~\cite{xia2020robust}, and Sel-CL+~\cite{li2022selective}; (3) Methods designed for tackling noisy labels on long-tailed data include HAR-DRW~\cite{cao2020heteroskedastic}, RoLT~\cite{wei2021robust}, and RoLT-DRW~\cite{wei2021robust}. The technical details of the above baselines are provided in Appendix~\textcolor{red}{C}. All experiments are run on NVIDIA Tesla V100 GPUs for fair comparisons.

\begin{table*}[!ht]
    \centering
    \setlength{\tabcolsep}{4mm}
    \renewcommand{\arraystretch}{1.1}
    \caption{Test accuracy (\%) on simulated CIFAR-10 and CIFAR-100 with varying noise rates and imbalance ratios. Note that all experiments are repeated five times. To avoid dense tables, we report the mean here. The best results are highlighted in \textcolor{red}{red}. The second best results are highlighted in \textcolor{blue}{blue}.}
    \resizebox{\textwidth}{!}{%
    \begin{tabular}{p{1cm}<{\centering}|c|ccccc|ccccc}
    \toprule[1.2pt]
    Dataset & Imbalance Ratio & \multicolumn{5}{c|}{10} & \multicolumn{5}{c}{100}\\
    \midrule
    \multirow{14}{*}{\rotatebox{90}{CIFAR-10}} & Noise Rate & 0.1 & 0.2 & 0.3 & 0.4 &0.5 &0.1 &0.2 & 0.3 & 0.4 & 0.5\\
     \cmidrule{2-12}
     ~ & ERM & 80.41 & 75.61 & 71.94 & 70.13 & 63.25 & 64.41 & 62.17 & 52.94 & 48.11 & 38.71  \\
    
    \cmidrule{2-12}
    ~ & LDAM & 84.59 & 82.37 & 77.48 & 71.41 & 60.30 & 71.46 & 66.26 & 58.34 & 46.64 & 36.66  \\
    ~ & LDAM-DRW & 85.94 & 83.73 & 80.20 & 74.87 & 67.93 &76.58 & 72.28 & 66.68 & 57.51 & 43.23  \\
    ~ & CRT & 80.22 & 76.15 & 74.17 & 70.05 & 64.15 & 61.54 & 59.52 & 54.05 & 50.12 & 36.73  \\
    ~ & NCM & 82.33 & 74.73 & 74.76 & 68.43 & 64.82 & 68.09 & 66.25 & 60.91 & 55.47 & 42.61  \\
    ~ & MiSLAS & \textcolor{blue}{87.58} & 85.21 & 83.39 & 76.16 & 72.46 & 75.62 & 71.48 & 67.90 & 62.04 & 54.54\\
    \cmidrule{2-12}
    ~ & Co-teaching & 80.30 & 78.54 & 68.71 & 57.10 & 46.77 & 55.58 & 50.29 & 38.01 & 30.75 & 22.85  \\
    ~ & CDR & 81.68 & 78.09 & 73.86 & 68.12 & 62.24 & 60.47 & 55.34 & 46.32 & 42.51 & 32.44 \\
    ~ & Sel-CL+ & 86.47 & 85.11 & \textcolor{blue}{84.41} & 80.35 & 77.27 & 72.31 & 71.02 & 65.70  & 61.37 & \textcolor{blue}{56.21} \\
    \cmidrule{2-12}
    ~ & HAR-DRW &  84.09 & 82.43 & 80.41 & 77.43 & 67.39 & 70.81 & 67.88 & 48.59 & 54.23 & 42.80  \\
    ~ & RoLT & 85.68 & 85.43 & 83.50 & 80.92 & 78.96 & 73.02 & 71.20 & 66.53 & 57.86 & 48.98 \\
    ~ & RoLT-DRW & 86.24 & \textcolor{blue}{85.49} & 84.11 & \textcolor{blue}{81.99} & \textcolor{blue}{80.05} & \textcolor{blue}{76.22} & \textcolor{blue}{74.92} & \textcolor{blue}{71.08} & \textcolor{blue}{63.61} & 55.06 \\
    \cmidrule{2-12}
    ~ & \textbf{RCAL (Ours)} & \textcolor{red}{88.09} & \textcolor{red}{86.46} & \textcolor{red}{84.58} & \textcolor{red}{83.43} & \textcolor{red}{80.80} & \textcolor{red}{78.60} & \textcolor{red}{75.81} & \textcolor{red}{72.76} & \textcolor{red}{69.78} & \textcolor{red}{65.05} \\
    \specialrule{1pt}{1pt}{5pt}
    \toprule[1pt]
    Dataset & Imbalance Ratio & \multicolumn{5}{c|}{10} & \multicolumn{5}{c}{100}\\
    \midrule
   \multirow{14}{*}{\rotatebox{90}{CIFAR-100}}  & Noise Rate & 0.1 & 0.2 & 0.3 & 0.4 &0.5 &0.1 &0.2 & 0.3 & 0.4 & 0.5\\
     \cmidrule{2-12}
    
    ~ & ERM & 48.54 & 43.27 & 37.43 & 32.94 & 26.24 & 31.81 & 26.21 & 21.79 & 17.91 & 14.23  \\
    
    \cmidrule{2-12}
    ~ & LDAM & 51.77 & 48.14 & 43.27 & 36.66 & 29.62 & 34.77 & 29.70 & 25.04 & 19.72 & 14.19  \\
    ~ & LDAM-DRW & 54.01 & 50.44 & 45.11 & 39.35 & 32.24 & 37.24 & 32.27 & 27.55 & 21.22 & 15.21  \\
    ~ & CRT & 49.13 & 42.56 & 37.80 & 32.18 & 25.55 & 32.25 & 26.31 & 21.48 & 20.62 & 16.01  \\
    ~ & NCM & 50.76 & 45.15 & 41.31 & 35.41 & 29.34 & 34.89 & 29.45 & 24.74 & 21.84 & 16.77  \\
    ~ & MiSLAS & \textcolor{red}{57.72} & \textcolor{blue}{53.67} & 50.04 & 46.05 & 40.63 & \textcolor{blue}{41.02} & \textcolor{blue}{37.40} & 32.84 & 26.95 & 21.84 \\
    \cmidrule{2-12}
    ~ & Co-teaching & 45.61 & 41.33 & 36.14 & 32.08 & 25.33 & 30.55 & 25.67 & 22.01 & 16.20 & 13.45  \\
    ~ & CDR & 47.02 & 40.64 & 35.37 & 30.93 & 24.91 & 27.20 & 25.46 & 21.98 & 17.33 & 13.64 \\
    ~ & Sel-CL+ & 55.68 & 53.52 & \textcolor{blue}{50.92} & \textcolor{blue}{47.57} & \textcolor{red}{44.86} & 37.45 & 36.79 & \textcolor{blue}{35.09} & \textcolor{blue}{31.96} & \textcolor{blue}{28.59}  \\
    \cmidrule{2-12}
    ~ & HAR-DRW &  51.04 & 46.24 & 41.23 & 37.35 & 31.30 & 33.21 & 26.29 & 22.57 & 18.98 & 14.78  \\
    ~ & RoLT & 54.11 & 51.00 & 47.42 & 44.63 & 38.64 & 35.21 & 30.97 & 27.60 & 24.73 & 20.14 \\
    ~ & RoLT-DRW & 55.37 & 52.41 & 49.31 & 46.34 & 40.88 & 37.60 & 32.68 & 30.22 & 26.58 & 21.05 \\
    \cmidrule{2-12} 
    ~ & \textbf{RCAL (Ours)} & \textcolor{blue}{57.50} & \textcolor{red}{54.85} & \textcolor{red}{51.66} & \textcolor{red}{48.91} & \textcolor{blue}{44.36} & \textcolor{red}{41.68} & \textcolor{red}{39.85} & \textcolor{red}{36.57} & \textcolor{red}{33.36} & \textcolor{red}{30.26} \\
    \bottomrule[1pt]
    \end{tabular}}
    \label{tab:CIFAR}
\end{table*}

\subsection{Datasets and Implementation Details}
\label{sec:4.2}
\myPara{Simulated noisy and class-imbalanced datasets.} We validate our method on CIFAR-10~\cite{krizhevsky2009learning} and CIFAR-100~\cite{krizhevsky2009learning} with varying noise rates and imbalance ratios. CIFAR-10 has 10 classes of images, including 50,000 training images and 10,000 testing images of size $32\times32$. CIFAR-100 also contains 50,000 training images and 10,000 testing images, but 100 classes. 

Specifically, to simulate realistic situations, we first create the imbalanced versions of CIFAR-10 and CIFAR-100 and then employ label noise injection. For the simulation of class-imbalanced datasets, we adopt long-tailed imbalance~\cite{cui2019class}. The long-tailed imbalance gradually reduces the number of examples in each class using an exponential function. In more detail, the exponential function is formulated as $n_k=n^o_kv^k$, where $n^o_k$ is the number of $k$-class examples in original datasets, $n_k$ is the number of $k$-class examples in long-tailed data, and $v\in(0,1)$. We consider
the most frequent classes occupying at least 50\% of the total training instances as head classes, and the remaining classes as tail classes. We employ the imbalance ratio $\rho$ to measure the imbalance degree, which is defined as the ratio between the sample size of the most frequent (head) class and that of the most scarce (tail) class. Additionally, for the generation of label noise,  we follow the setting of RoLT~\cite{wei2021robust}. Let $T_{ij}(\xx)$ be the probability that the true label $i$ is corrupted to the noisy label $j$ for instance $\xx$. The label flipping process is correlated with the number of each class. Given the noise rate $\eta$, we define: $T_{ij}(\xx)=\mathbbm{P}[\tilde{Y}=j|Y=i,\xx]=1-\eta$ if $i=j$ and otherwise $T_{ij}(\xx)=\mathbbm{P}[\tilde{Y}=j|Y=i,\xx]=\frac{n_j}{n-n_i}\eta$,
where $Y$ and $\tilde{Y}$ denote the random variables of clean labels and noisy labels, respectively. In the following experiments, the imbalanced
ratio $\rho$ is chosen in $\{10,100\}$. The noise rate is $\eta$ is chosen in $\{0.1,0.2,0.3,0.4,0.5\}$. 

For both CIFAR-10 and CIFAR-100 datasets, we use a ResNet-32~\cite{he2016deep} network. We perform the strong augmentations SimAug~\cite{chen2020simple} in the contrastive learning stage and standard weak augmentations in the classifier learning stage. In the contrastive learning stage, we employ the official MOCO implementation in PyTorch\footnote{\url{https://github.com/facebookresearch/moco.git}}. The model of contrastive learning is trained for $1000$ epochs in total, and the queue size is set to $4096$. In the classifier learning stage, the batch size is $128$, and we run $100$ epochs for CIFAR-10 and $200$ epochs for CIFAR-100. The number of selected head classes, \textit{i.e.}, $q$, is set to 3. Since the sampled data points are deep representations before the linear layer, we adopt twp SGD optimizers (momentum $0.9$) for datasets reduced from original datasets and datasets built by sampled data points. For the former, we give different initial learning rates to the backbone and linear head, which are set to $0.01$ and $1$. We reduce them by a factor of $10$ at $\{20, 40, 60, 80\}$-th epoch. For the latter, the learning rate is set to 0.001.

\myPara{Real-world noisy and imbalanced datasets.} We also evaluate RCAL on real-world datasets, \textit{i.e.,} WebVision~\cite{li2017webvision} and Clothing1M~\cite{xiao2015learning}.   WebVision contains 2.4 million images crawled from the website using the 1,000 concepts shared with ImageNet ILSVRC12. Following the ``mini'' setting in~\cite{ma2020normalized,chen2019understanding}, we take the first 50 classes of the Google resized image subset and name it WebVision-50. We then test the trained network on the same 50 classes of the WebVision validation set and ILSVRC12 validation set. For WebVision-50, we use an Inception-ResNet-v2 network and train it using SGD with a momentum of 0.9, a weight decay of $10^{-4}$, and a batch size of 64. Clothing1M contains 1 million training images, and 50k, 14k, 10k images with clean labels for training, validating and testing, but with
14 classes. Note that we do not use the 50k and 14k clean data in experiments, since it is more practical that
there is no extra clean data. We exploit a ResNet-50 network for Clothing1M. The optimizer is Adam with a learning rate of 0.001 and a batch size of 256.

\subsection{Results on Simulated CIFAR-10/100}
Results on simulated CIFAR-10 and CIFAR-100 are shown in Table~\ref{tab:CIFAR}. We analyze the results as follows. We observe that RCAL can outperform all baselines under almost all noise rates on both simulated CIFAR-10 and CIFAR-100. Compared to ERM, RCAL averagely gains over $11\%$ and $13\%$ accuracy improvements on simulated CIFAR-10 and CIFAR-100, respectively. As the task being more challenging, RCAL exhibits a more distinct improvement. Particularly, for CIFAR-10, RCAL can achieve over $8\%$ higher test accuracy than the second best baseline Sel-CL+, in the case of the imbalance ratio $100$ and the noise rate $0.5$. Moreover, some of the baselines' performances are inferior to the ERM, \textit{e.g.,} Co-teaching. Co-teaching employs a small loss trick to identify potential clean data. However, examples of tail classes tend to have larger training losses which makes it hard to be selected for training whether the labels are clean or not. This leads to a more extreme imbalanced data distribution, which degenerates performance.  

\subsection{Results on Real-world Noisy and Imbalanced Datasets}
Table~\ref{tab:webvision} shows the results on WebVision-50. As can be seen, RCAL achieves the best results on top-5 accuracy on both the WebVision validation set and ImageNet ILSVRC12 validation set compared to other state-of-the-art methods. As Sel-CL+ uses a ResNet-18 network for WebVision-50, we do not include this method for comparison. Note that the competitive baseline RoLT+, based on RoLT, employs semi-supervised learning techniques to boost performance. Therefore, for a fair comparison, we also combine semi-supervised learning algorithms to boost RCAL (referred as RCAL+). Moreover, Table~\ref{tab:clothing1m} shows the results on Clothing1M. We observe that RCAL+ achieves state-of-the-art performance, which verifies the effectiveness of our proposed method against real-world noisy and imbalanced datasets.
\begin{table}[t]
    \centering
    \setlength{\tabcolsep}{2mm}
    \renewcommand{\arraystretch}{1.1}
    \caption{Top1 and Top5 test accuracy on Webvision and ImageNet validation sets. Partial
numerical results come from~\cite{wei2021robust,cao2020heteroskedastic}. The best results are in \textbf{bold}. } 
\resizebox{\linewidth}{!}{%
    \begin{tabular}{lcccc}
    \toprule
    Train & \multicolumn{4}{c}{WebVision-50}\\
    Test  & \multicolumn{2}{c}{WebVision} & \multicolumn{2}{c}{ILSVRC12}  \\
    
    Method & Top1 (\%) & Top5 (\%) & Top1 (\%) & Top5 (\%)  \\
    \midrule
    ERM & 62.5 & 80.8 & 58.5 & 81.8\\
    Co-teaching~\cite{han2018co} & 63.58 & 85.20 & 61.48 & 84.70\\
    INCV~\cite{chen2019understanding} & 65.24 & 85.34 & 61.60 & 84.98\\
    MentorNet~\cite{jiang2018mentornet} & 63.00 & 81.40 & 57.80 & 79.92 \\
    CDR~\cite{xia2020robust} & - & -  & 61.85 & - \\
    HAR~\cite{cao2020heteroskedastic} & 75.5 & 90.7 & 70.3 & 90.0\\
    RoLT+~\cite{wei2021robust} & 77.64 & 92.44 & 74.64 & 92.48 \\
    \midrule
    RCAL (Ours) & 76.24 & 92.83 & 73.60 & 93.16 \\
    \textbf{RCAL+ (Ours)} & \textbf{79.56} & \textbf{93.36} & \textbf{76.32} & \textbf{93.68}\\
    \bottomrule
    \end{tabular}}
	\label{tab:webvision}

\end{table}

\begin{table}[t]
    \centering
    \setlength{\tabcolsep}{3mm}
    \renewcommand{\arraystretch}{1.1}
    \caption{Test accuracy on the Clothing1M test dataset. Partial numerical results come from~\cite{zhang2021learning}. The best results are in \textbf{bold}.} 
    \resizebox{\linewidth}{!}{%
    \begin{tabular}{lc|lc}
    \toprule
    Method & Top1 (\%) & Method & Top1 (\%) \\
    \midrule
    ERM & 68.94 & Co-teaching~\cite{han2018co} & 67.94\\
    MentorNet~\cite{jiang2018mentornet} &  67.25 & CDR~\cite{xia2020robust} & 68.25\\
    Forward~\cite{patrini2017making} & 69.84 &  D2L~\cite{ma2018dimensionality}  & 69.74\\
    Joint~\cite{tanaka2018joint} & 72.23 & GCE~\cite{zhang2018generalized} & 69.75 \\
    Pencil~\cite{yi2019probabilistic} & 73.49 & LRT~\cite{zheng2020error} & 71.74 \\
    SL~\cite{wang2019symmetric} & 71.02 & MLNT~\cite{li2019learning} & 73.47 \\
    PLC~\cite{zhang2021learning}& 74.02 &
    DivideMix~\cite{li2020dividemix} & 74.76 \\
    ELR+~\cite{liu2020early} & 74.81 & \textbf{RCAL+ (Ours)} & \textbf{74.97}\\
    \bottomrule
    \end{tabular}}
	\label{tab:clothing1m}
\end{table}

\begin{table}[!t]
    \centering
    \setlength{\tabcolsep}{2mm}
    \renewcommand{\arraystretch}{1.1}
    \caption{Ablation study of test accuracy (\%) on simulated CIFAR-10. We report the mean of five trials. The best results are in \textbf{bold}. ``CL'' means unsupervised contrastive learning. ``DC'' means distributional calibration. ``REG'' means individual calibration by restricting the distance between subsequent learned representations and the representations brought by unsupervised contrastive learning. }
    \resizebox{\linewidth}{!}{%
    \begin{tabular}{l|cc|cc}
    \toprule
    Dataset & \multicolumn{4}{c}{CIFAR-10} \\
    \midrule
    Imbalance Ratio & \multicolumn{2}{c|}{10} & \multicolumn{2}{c}{100}  \\
    \midrule
    Noise Rate & 0.2 & 0.4 & 0.2 & 0.4 \\
    \midrule
    RCAL  &  \textbf{86.46} & \textbf{83.43} & \textbf{75.81} & \textbf{69.78}\\
    RCAL w/o Mixup & 84.08 & 79.27 & 72.47 & 64.83 \\
    RCAL w/o Mixup, REG & 83.23 & 78.12 & 67.49 & 58.27 \\
    RCAL w/o Mixup, REG, DC & 80.40 & 74.37 & 64.02 & 54.61  \\
    RCAL w/o Mixup, REG, DC, CL & 75.61 & 70.13 & 62.17 & 48.11 \\
    
    \bottomrule
    \end{tabular}}
    \label{tab:components1}
\end{table}

\subsection{Ablation Study}

\myPara{Impact of each component.} To explore what makes RCAL successful, we report the test accuracy on simulated CIFAR-10 by removing each component gradually. Table~\ref{tab:components1} shows the contribution of each component to our method. The experiments on simulated CIFAR-100 can be checked in Appendix~\textcolor{red}{B.1}.

\myPara{The influence of batch sizes.} To study the impact of batch size. We provide results
with batch sizes 64, 128, and 256 respectively, which are
shown in Table~\ref{tab:batch_size}. As can be seen, our RCAL is overall stable to the change of batch sizes in a certain range.

\begin{table}[!t]
    \centering
    \setlength{\tabcolsep}{4mm}
    \renewcommand{\arraystretch}{1.1}
    \caption{Test accuracy (\%) of RCAL with different batch sizes.}
    \resizebox{\linewidth}{!}{%
    \begin{tabular}{c|cc|cc}
    \toprule
        Dataset & \multicolumn{2}{c|}
        {CIFAR-10} & \multicolumn{2}{c}{CIFAR-100}  \\
        \midrule
        Noise Rate & \multicolumn{2}{c|}{0.5} & \multicolumn{2}{c}{0.5} \\
        \midrule
        Imbalance Ratio & 10 & 100 & 10 & 100\\
        \midrule
        Batch size of 64 & 79.53 & 65.69 & 44.15 & 29.93 \\
        Batch size of 128 & 80.80 & 65.05 & 44.36 & 30.26 \\
        Batch size of 256 & 80.92 & 63.27 & 43.22 & 29.44 \\
        \bottomrule
    \end{tabular}}
    \label{tab:batch_size}
\end{table}

\myPara{Fine-grained results and analysis.} To further analyze how RCAL affects classes with different sizes, we divide classes into three splits according to the state-of-the-art method~\cite{zhong2021improving}: \textit{Many}, \textit{Medium}, and \textit{Few} classes. We report classification performance on test data of the three splits in Table~\ref{tab:medium}. As can be seen, compared with ERM, both MiSLAS can improve the performance on \textit{Medium} and \textit{Few} classes, leading to final better overall performance. However, compared with our RCAL, MiSLAS overemphasizes the model performance on \textit{Few} classes, but somewhat ignores the performance on \textit{Many} and \textit{Medium} that also are important. Therefore, as for overall performance, our RCAL surpasses MiSLAS by a clear margin. 

\myPara{Sensitivity analysis of hyper-parameters.} We analyze the sensitivity of hyper-parameters to value changes. Here, different network structures, \ie, ResNet-32, ResNet-18, and ResNet-34, are employed. The value of the disturbance degree $\alpha$ is chosen in the range $\{0.001, 0.01, 0.1, 1\}$, while the value of the regularization strength $\beta$ is chosen in the range $\{0.1,0.2,0.3,0.4,0.5\}$. We report the analysis results in Appendix~\textcolor{red}{B.2}. With different network structures, the achieved performance by our method is stable with the changes of hyper-parameters. The advantage makes it easy to apply our method in practice.

\begin{table}[!t]
    \centering
    \setlength{\tabcolsep}{4.5mm}
    \renewcommand{\arraystretch}{1.1}
    \caption{Test accuracy (\%) of many/medium/few classes on CIFAR-10, where the noise rate and imbalance ratio are 0.5 and 10.}
    \resizebox{\linewidth}{!}{%
    \begin{tabular}{c|cccc}
    \toprule
        Method & Many & Medium & Few & Overall \\
        \midrule
        ERM & 82.71 & 55.31 & 57.22 & 63.25 \\
        MiSLAS & 67.16 & 69.52 & 81.66 & 72.46 \\
        RCAL (Ours) & 84.10 & 84.13 & 73.98 & 80.80\\
        \bottomrule
    \end{tabular}}
    \label{tab:medium}
\end{table}

\section{Conclusion}
\label{sec:conclusion}
This paper proposes a representation calibration method (RCAL) to handle a realistic while challenging problem: learning with noisy labels on long-tailed data. We suppose that before training data are corrupted and imbalanced, the deep representations of instances in each class conform to a multivariate Gaussian distribution. Using the representations learned from unsupervised contrastive learning, we recover the underlying representation distributions and then sample data points to balance the classifier. In classifier training, we further take advantage of representation robustness brought by contrastive learning to improve the classifier's performance. Extensive experiments demonstrate that our methods can help improve the robustness against noisy labels and long-tailed data simultaneously. In the future, we are interested in adapting our method to other domains, such as natural language processing and speech recognition. Furthermore, we are also interested in exploring the possibilities of using other multivariate-distribution assumptions on deep representations and deriving theoretical results based on them, \textit{e.g.}, the Laplace distribution~\cite{eltoft2006multivariate}, Sub-Gaussian distribution~\cite{devroye2016sub}, and Cauchy distribution~\cite{arnold2000skew}.

\subsection*{Acknowledgements}

\tolerance=2000
Chun Yuan was supported by the National Key R\&D Program of China (2022\allowbreak YFB\allowbreak 4701400\allowbreak /\allowbreak 4701402) and SZSTC Grant (JCYJ\allowbreak 2019\allowbreak 0809172201639, WDZC\allowbreak 2020\allowbreak 0820200655001).

Weiran Huang is partially funded by Microsoft Research Asia.

{\small
\bibliographystyle{ieee_fullname}
\bibliography{egbib}
}

\clearpage
\appendix
\onecolumn
\setlength{\parindent}{0pt}
\begin{center}
    \Large\bf Appendix
\end{center}
\vspace{0.4cm}
\section{Proof of Theoretical Results}

\begin{theorem}
Under Assumption \ref{assumption}, there exists constant $C$ such that
\begin{equation*}
\mathbb{E}\|\hat{\bm{\mu}}_k - \bm{\mu}_k\|^2 \leq C\cdot\left[ \eta^2 + \frac{m}{n_{\text{tail}}}\right]
\end{equation*}
and
\begin{equation*}
\mathbb{E}\|\hat{\bm{\mu}}'_k - \bm{\mu}_k\|^2 
\leq C\cdot\left[ \eta^2 + \delta_q^2 + \max\{\frac{1}{q\gamma}\cdot\frac{m}{{n}_{\text{tail}}}, \frac{m}{{n}_{\text{head}}}\}\right].
\end{equation*}
\end{theorem}
\begin{proof} 

After some calculation,

\begin{equation*}
    \mathbb{P}(y=j\,|\, \tilde{y}=k) = \frac{\mathbb{P}(y = j)\mathbb{P}(\tilde{y}=k\,|\, y=j)}{\mathbb{P}(\tilde{y}=k)} = P(\tilde{y}=j\,|\,y =k).
\end{equation*}
Therefore, 
\begin{equation*}
    \mathbb{E}\hat{\bm{\mu}}_k = \mathbb{E}[\bm{z}\,|\,\tilde{y}=k] = \sum_{j=1}^K \mathbb{P}(y=j\,|\,\tilde{y}=k) \bm{\mu}_j = (1-\eta+\frac{n_k}{n}\eta)\bm{\mu}_k + \sum_{j\not=k} \frac{n_j \eta}{n}\bm{\mu}_j.
\end{equation*}

\begin{equation*}
\begin{aligned}
\mathbb{E} \|\hat{\bm{\mu}}_k-\bm{\mu}_k\|^2 &= \mathbb{E} \|\hat{\bm{\mu}}_k-\mathbb{E}\hat{\bm{\mu}}_k\|^2 + \|\mathbb{E}\hat{\bm{\mu}}_k - \bm{\mu}_k\|^2 \\
&\leq \|\sum_{j\not=k} \frac{n_j\eta}{n}(\bm{\mu}_j - \bm{\mu}_k)\|^2 + \frac{m\sigma^2}{n} \\
&\leq C(\eta^2 + \frac{m}{n}). 
\end{aligned}
\end{equation*}

The calibrated mean can be written as
\begin{equation*}
    \hat{\bm{\mu}}'_k = \sum_{j\in\mathcal{C}_k^q,j\not=k} \frac{\tau}{1+(q-1)\tau} \hat{\bm{\mu}}_j + \frac{1}{1+(q-1)\tau}\hat{\bm{\mu}}_k.
\end{equation*}

Therefore,
\begin{equation*}
\begin{aligned}
\mathbb{E} \|\hat{\bm{\mu}}'_k-\bm{\mu}_k\|^2 &= \mathbb{E} \|\hat{\bm{\mu}}'_k-\mathbb{E}\hat{\bm{\mu}}'_k\|^2 + \|\mathbb{E}\hat{\bm{\mu}}'_k - \bm{\mu}_k\|^2 \\
&\leq C\eta^2 + \delta_q^2 + \frac{1}{q\tau}\cdot \frac{m}{n_k}.
\end{aligned}
\end{equation*}
This finishes the proof.
\end{proof}
\vspace{0.2cm}
\section{Detailed Results for Ablation Study}

\subsection{The Impact of Each Component}
Table~\ref{tab:components} shows the detailed results of RCAL on simulated 
 CIFAR-10 and CIFAR-100 with the imbalance ratio $\rho = \{10, 100\}$ and the noise rate $\eta = \{0.2, 0.4\}$. We observe that, constrastive learning can largely enhance the deep representations, and the followed distributional calibration further improves the classification performance, which justifies our claims.
 
\begin{table}[!h]
    \centering
    \setlength{\tabcolsep}{4.5mm}
    \renewcommand{\arraystretch}{1.1}
    \caption{Ablation study results of test accuracy (\%) on simulated CIFAR-10 and CIFAR-100. We report the mean. The best results are in \textbf{bold}. In the following, ``CL'' means unsupervised contrastive learning. ``DC'' means distributional calibration. ``REG'' means individual calibration by restricting the distance between subsequently learned representations and the representations brought by unsupervised contrastive learning. }
    \resizebox{\linewidth}{!}{%
    \begin{tabular}{l|cc|cc|cc|cc}
    \toprule
    Dataset & \multicolumn{4}{c|}{CIFAR-10} & \multicolumn{4}{c}{CIFAR-100}\\
    \midrule
    Imbalance Ratio & \multicolumn{2}{c|}{10} & \multicolumn{2}{c|}{100} & \multicolumn{2}{c|}{10} & \multicolumn{2}{c}{100} \\
    \midrule
    Noise Rate & 0.2 & 0.4 & 0.2 & 0.4 & 0.2 & 0.4 & 0.2 & 0.4  \\
    \midrule
    RCAL  &  \textbf{86.46} & \textbf{83.43} & \textbf{75.81} & \textbf{69.78} & \textbf{54.85} & \textbf{48.91} & \textbf{39.85} & \textbf{33.36}\\
    RCAL w/o Mixup & 84.08 & 79.27 & 72.47 & 64.83 & 51.22 & 45.53 & 36.78 & 30.85\\
    RCAL w/o Mixup, REG & 83.23 & 78.12 & 67.49 & 58.27 & 48.74 & 42.15 & 34 31 &27.14\\
    RCAL w/o Mixup, REG, DC & 80.40 & 74.37 & 64.02 & 54.61 & 47.01 & 40.85 & 32.27 & 25.42 \\
    RCAL w/o Mixup, REG, DC, CL & 75.61 & 70.13 & 62.17 & 48.11 & 43.27 & 32.94 & 26.21 & 17.91\\
    \bottomrule
    \end{tabular}}
    \label{tab:components}
\end{table}

\subsection{Sensitivity Analysis of Hyper-Parameters}
\begin{figure}[!h]
\centering
\subfloat[]{\includegraphics[width=1.6in]{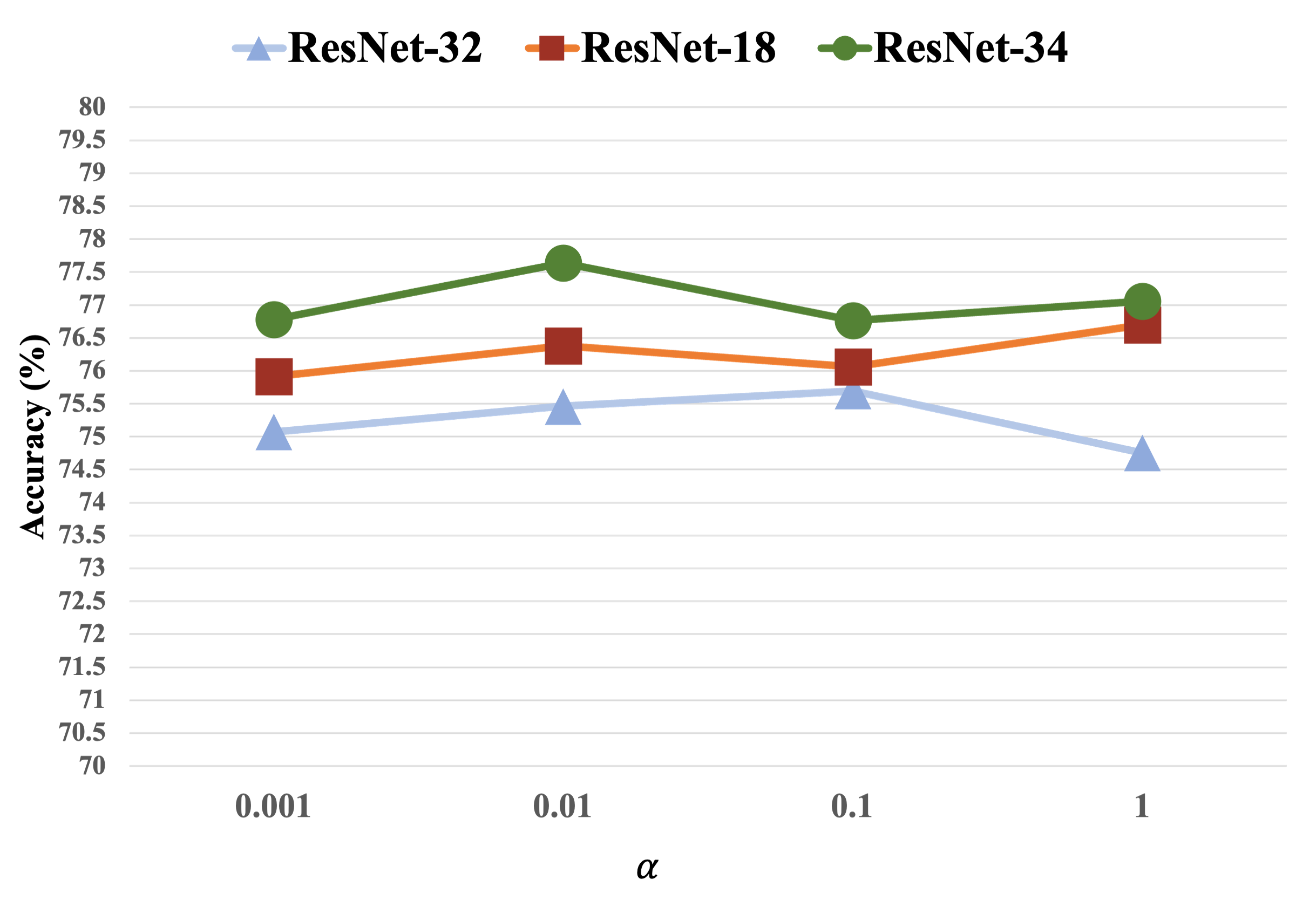}}
\subfloat[]{\includegraphics[width=1.6in]{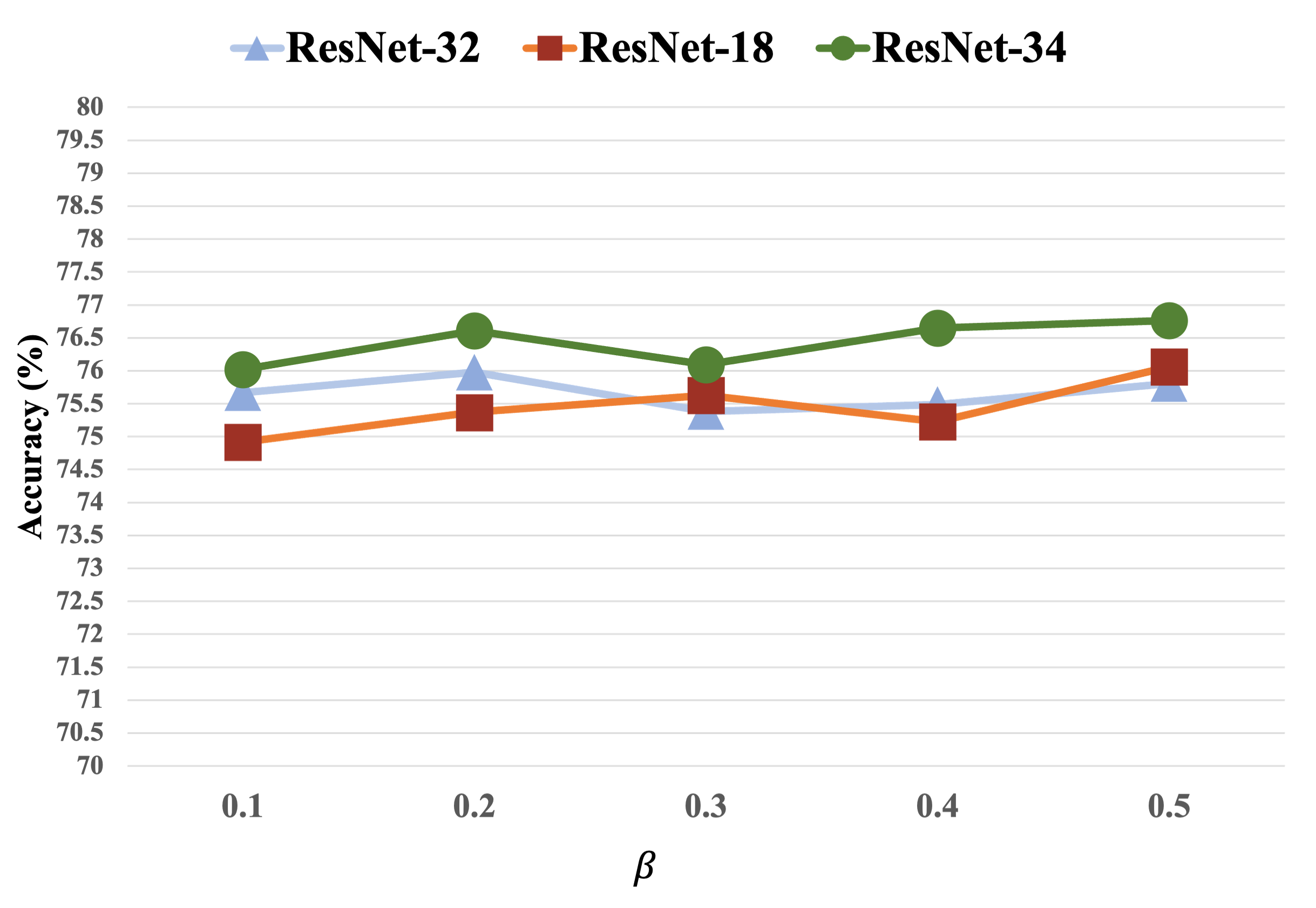}}
\quad
\subfloat[]{\includegraphics[width=1.6in]{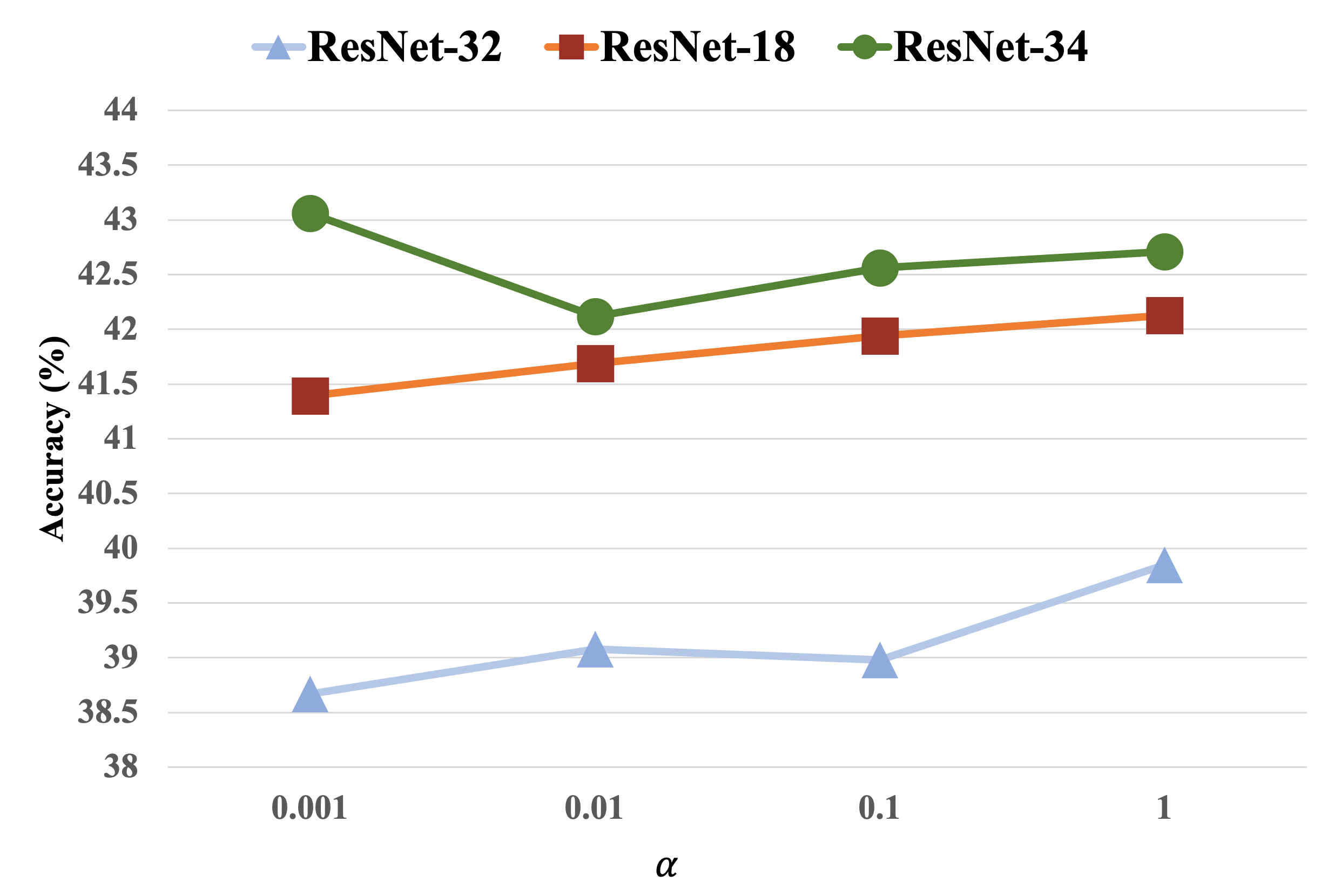}}
\subfloat[]{\includegraphics[width=1.6in]{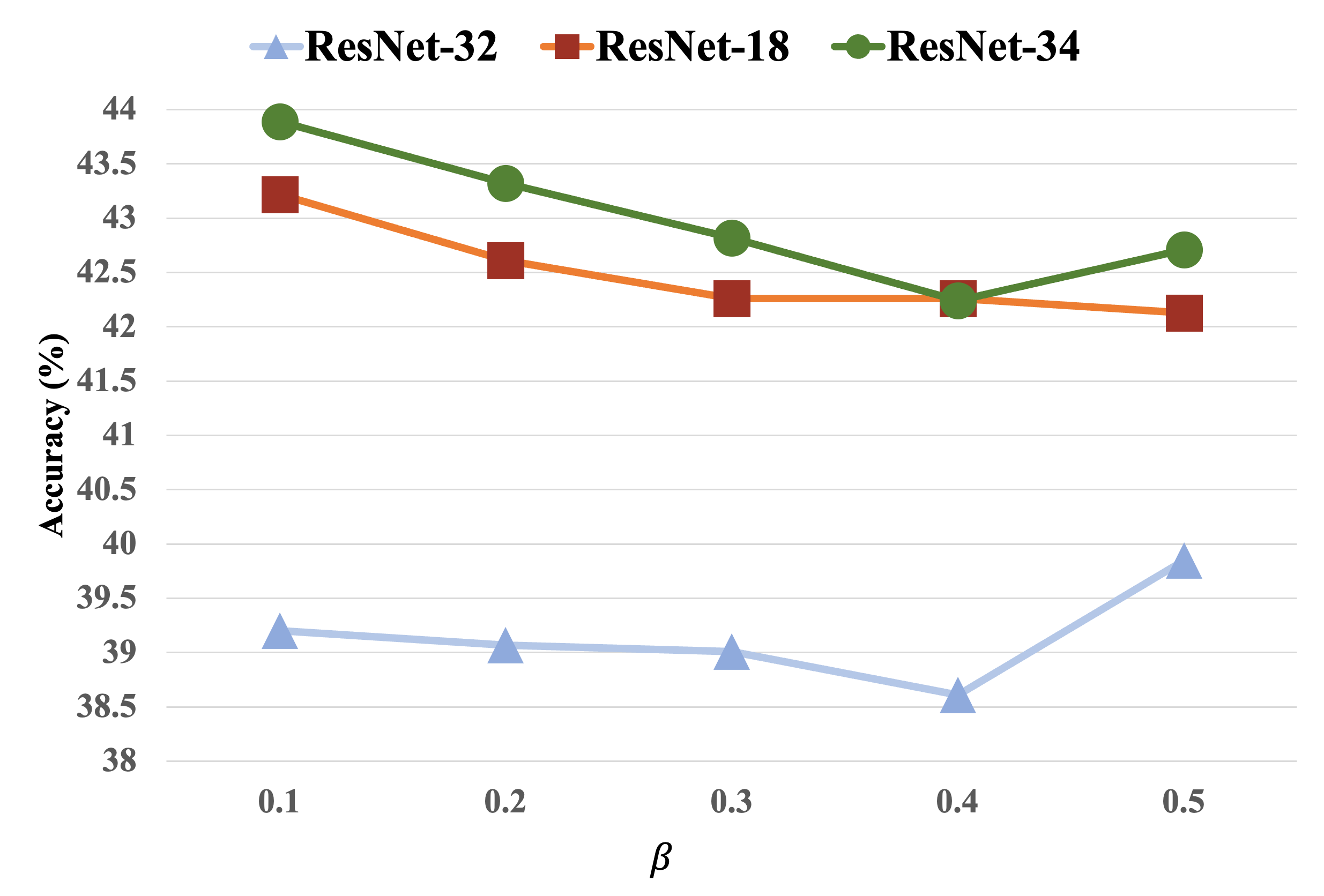}}
\caption{The influence of hyper-parameters with different values on simulated CIFAR-10 and CIFAR-100 under the imbalance ratio $100$ and noise rate $0.2$. Subfigures (a) and (c) present the results on noisy and imbalanced CIFAR-10 and CIFAR-100 under different values of $\alpha$. Additionally, subfigures (b) and (d) show the results on noisy and imbalanced CIFAR-10 and CIFAR-100 under different values of $\beta$.}
\label{abla:para}
\end{figure}

We explore the influence of hyper-parameters with different values in Figure~\ref{abla:para}. It can be seen that RCAL is not sensitive to the changes of hyper-parameters.

\subsection{Representation Visualizations}
\begin{figure*}[!tp]
\begin{minipage}[c]{0.05\columnwidth}~\end{minipage}%
    \begin{minipage}[c]{0.235\textwidth}\centering\small -- \textbf{\footnotesize{Training  ($\rho=100,\eta=0.2$)}} -- \end{minipage}%
    \begin{minipage}[c]{0.235\textwidth}\centering\small -- \textbf{\footnotesize{Test  ($\rho=100,\eta=0.2$)}} -- 
    \end{minipage}%
    \begin{minipage}[c]{0.235\textwidth}\centering\small -- \textbf{\footnotesize{Training  ($\rho=100,\eta=0.4$)}} -- 
    \end{minipage}
    \begin{minipage}[c]
    {0.235\textwidth}\centering\small -- \textbf{{\footnotesize{Test ($\rho=100,\eta=0.4$)}}} -- 
    \end{minipage}
    \begin{minipage}[c]{0.05\columnwidth}\centering\small \rotatebox[origin=c]{90}{-- ERM --} \end{minipage}%
\begin{minipage}[c]{0.95\textwidth}
        \includegraphics[width=0.25\textwidth]{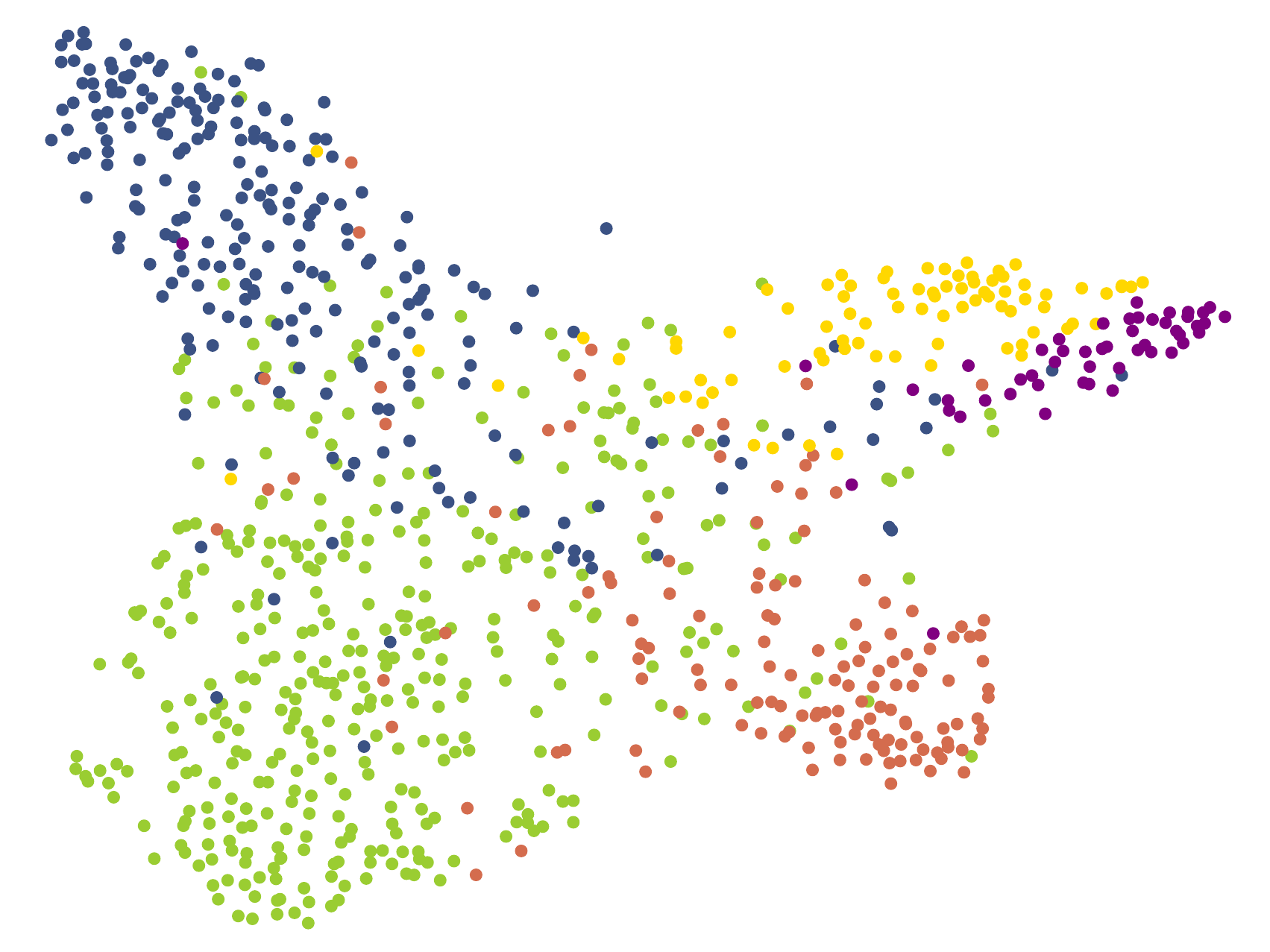}%
        \includegraphics[width=0.25\textwidth]{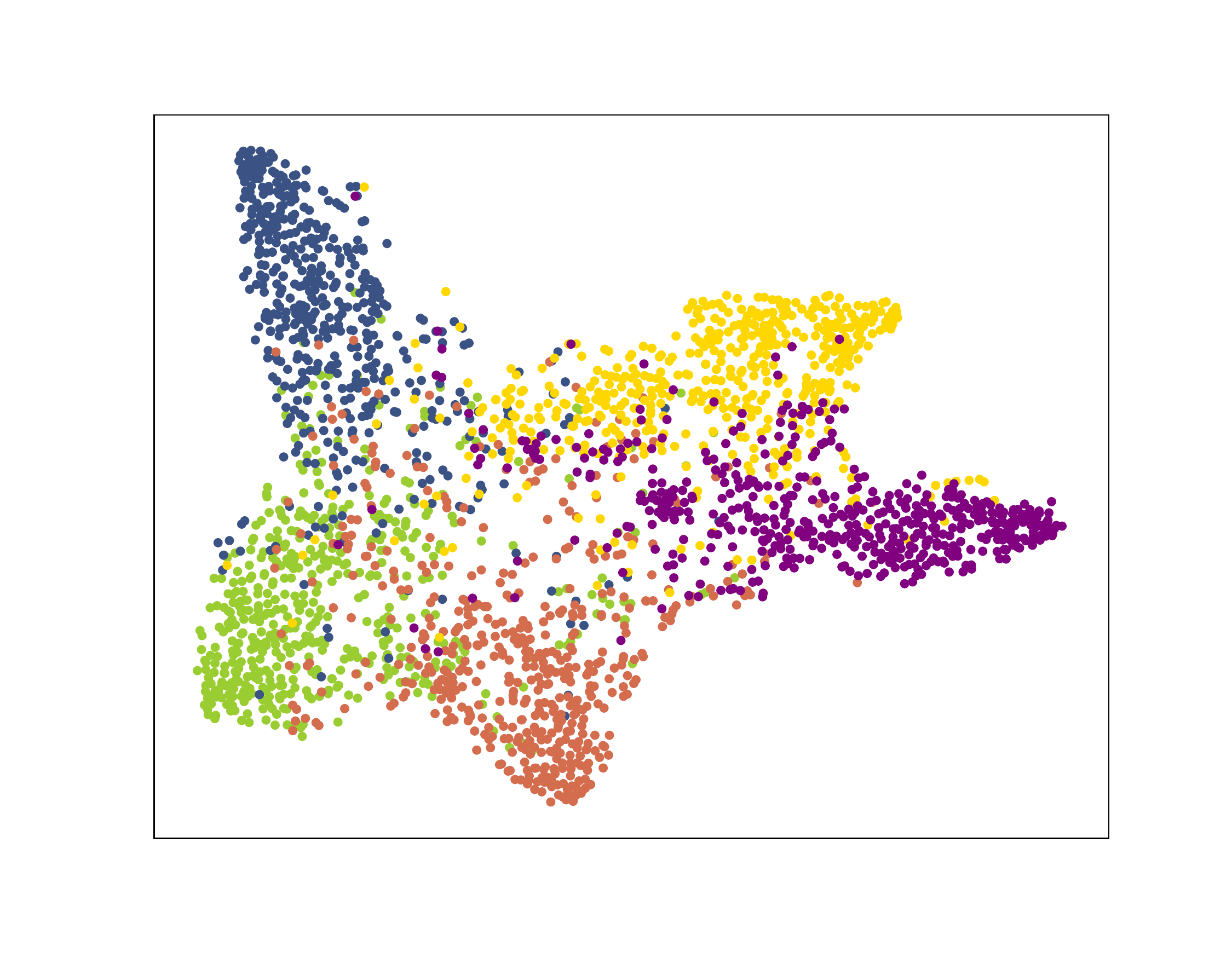}%
        \includegraphics[width=0.25\textwidth]{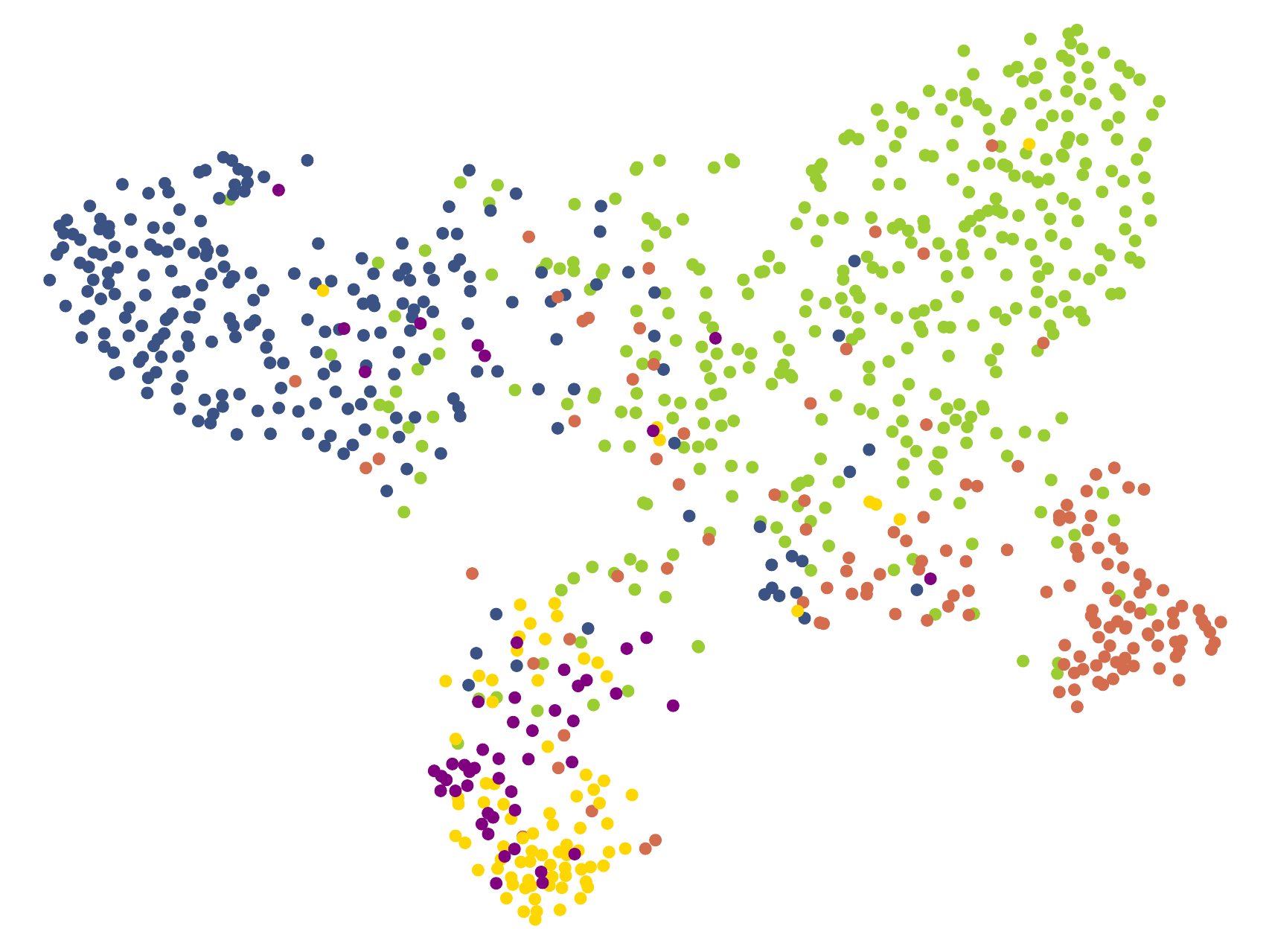}%
        \includegraphics[width=0.25\textwidth]{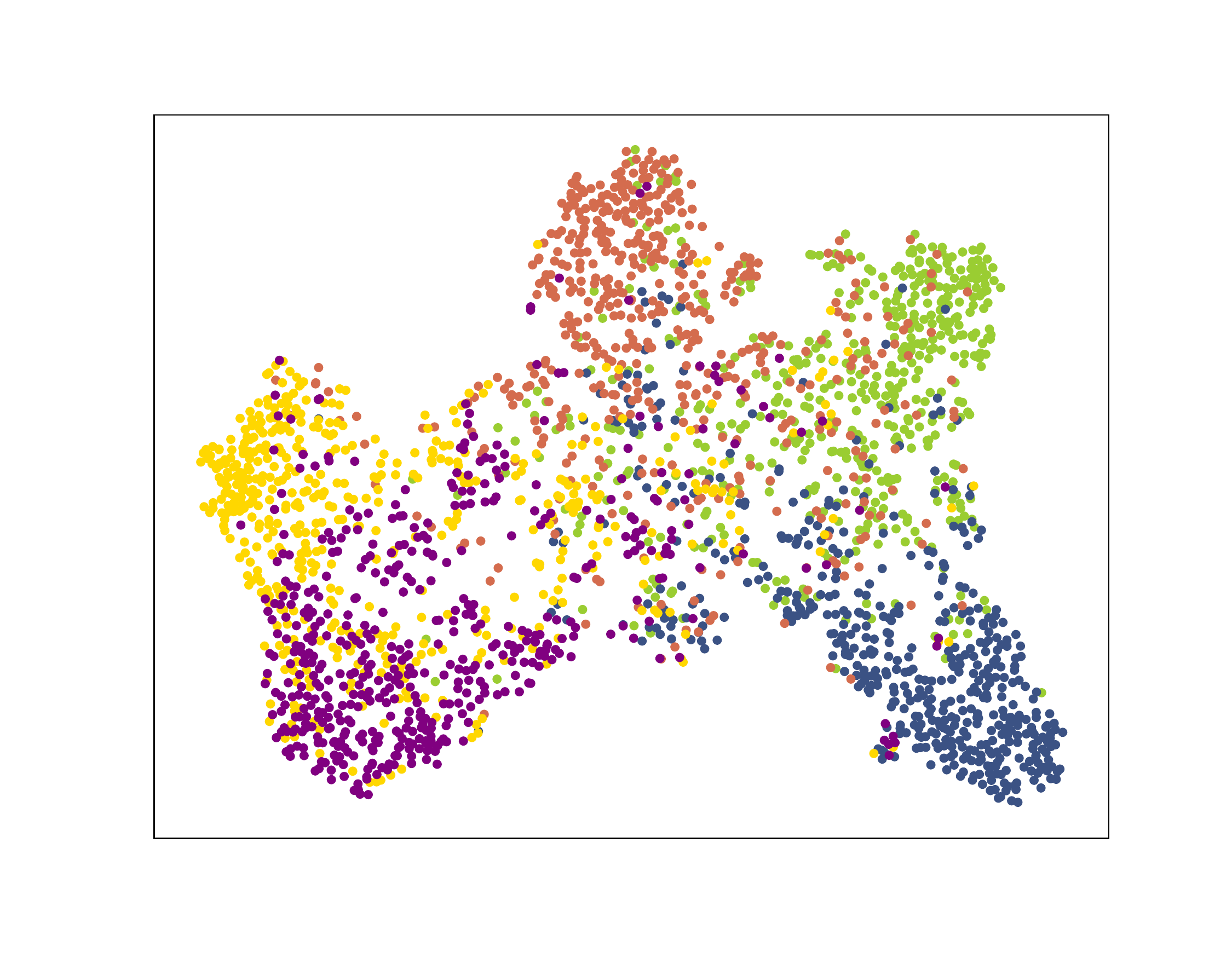}%
    \end{minipage}
    \begin{minipage}[c]{0.05\columnwidth}\centering\small \rotatebox[origin=c]{90}{-- RCAL --} \end{minipage}%
\begin{minipage}[c]{0.95\textwidth}
        \includegraphics[width=0.25\textwidth]{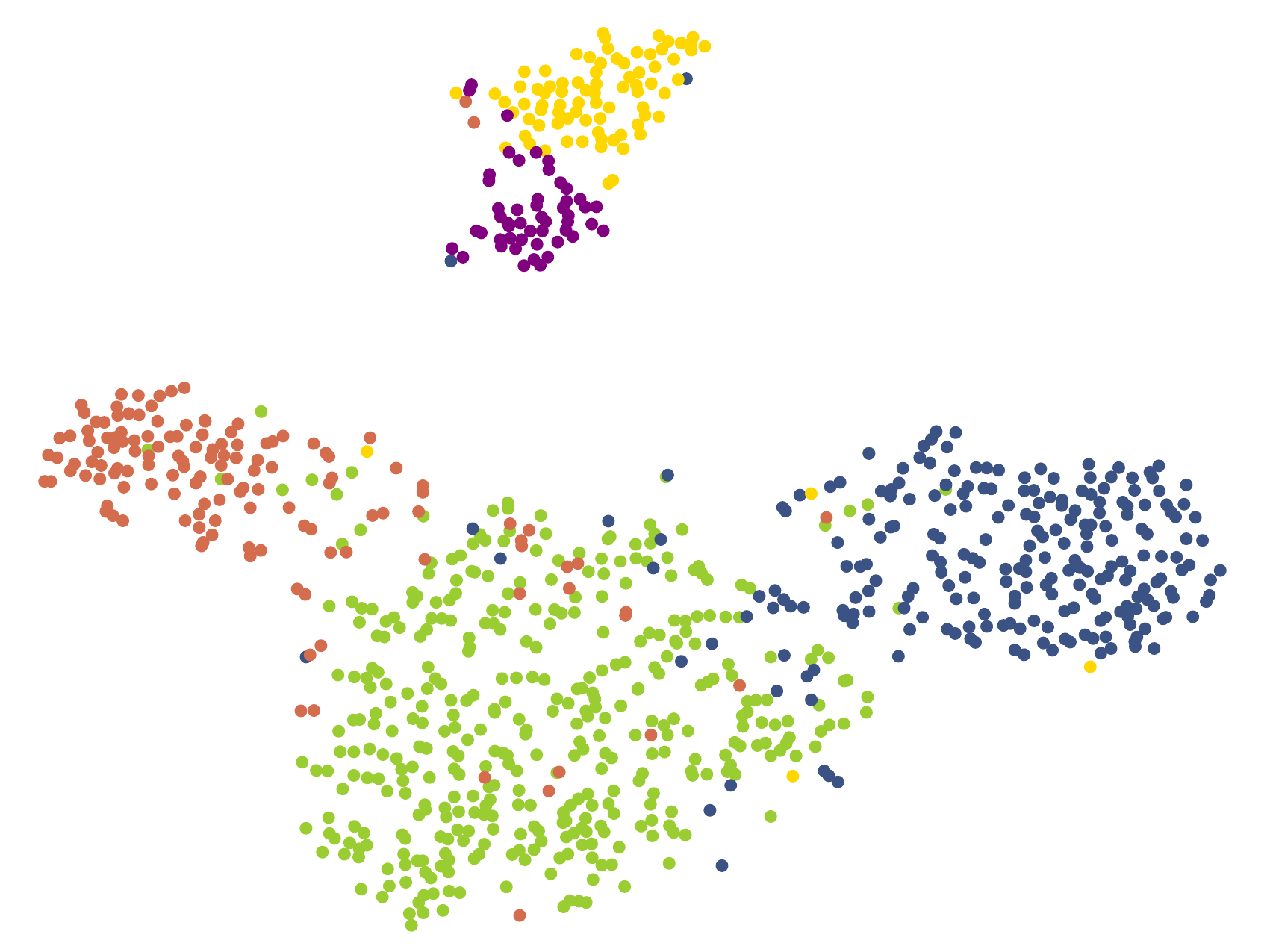}%
        \includegraphics[width=0.25\textwidth]{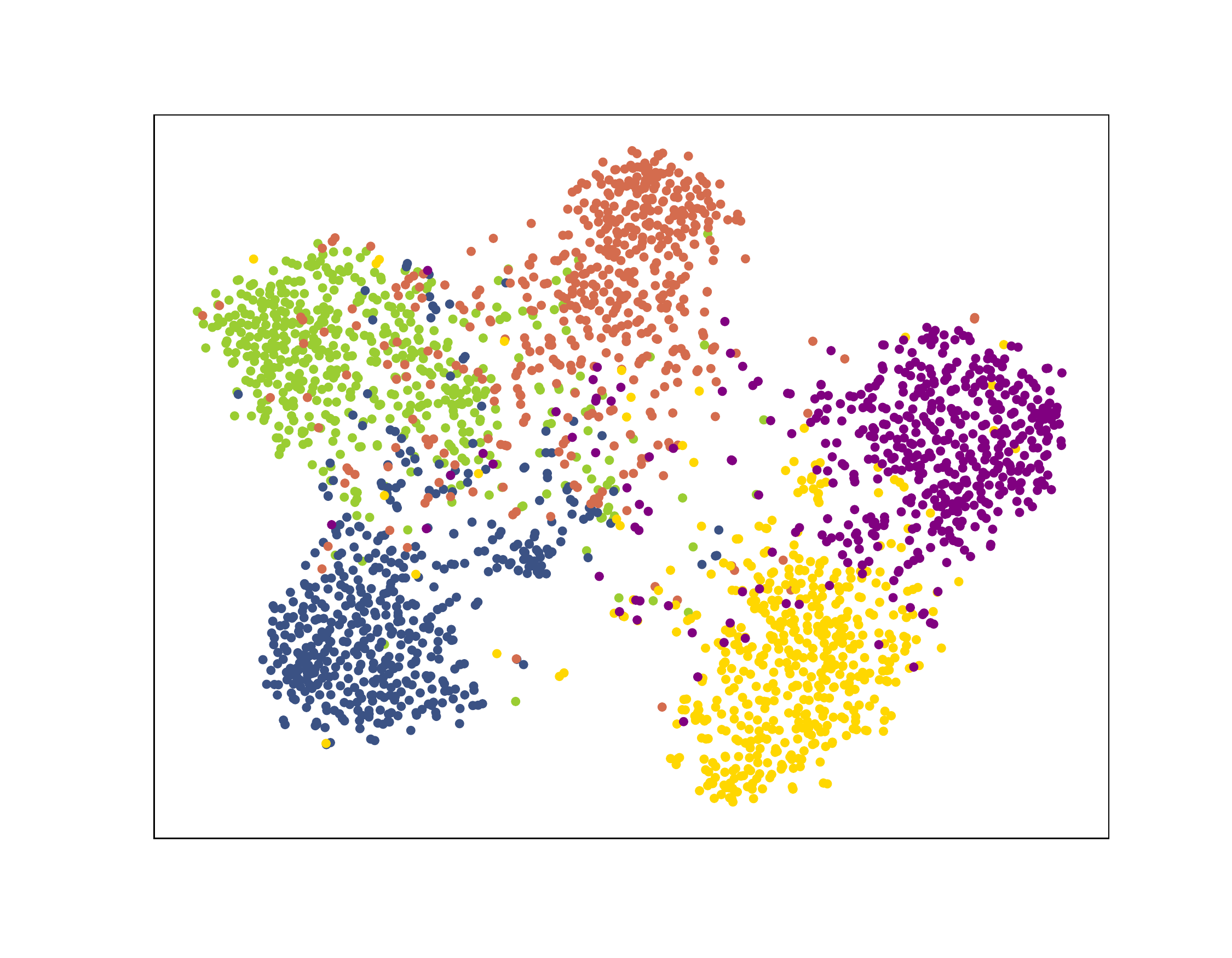}%
        \includegraphics[width=0.25\textwidth]{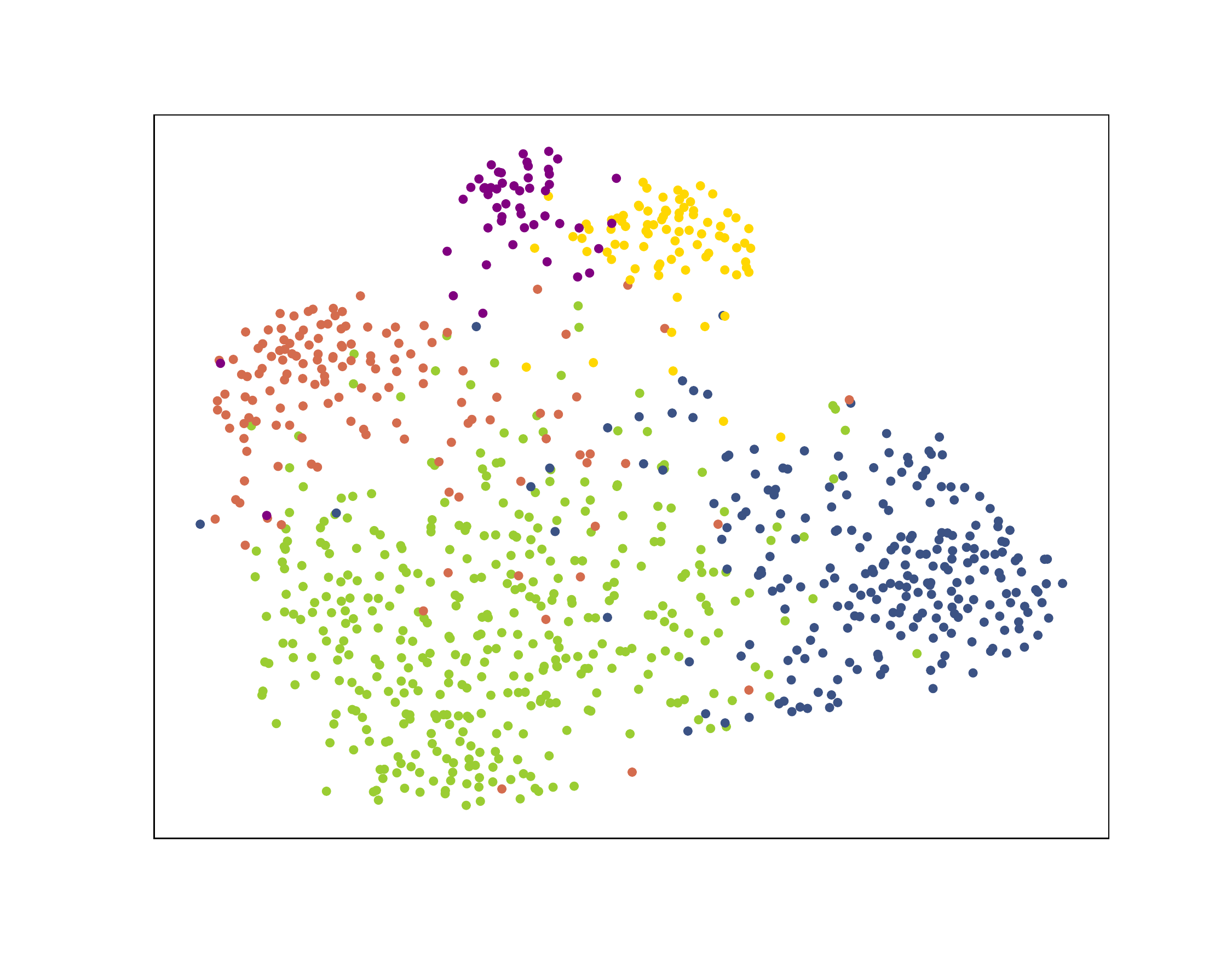}%
        \includegraphics[width=0.25\textwidth]{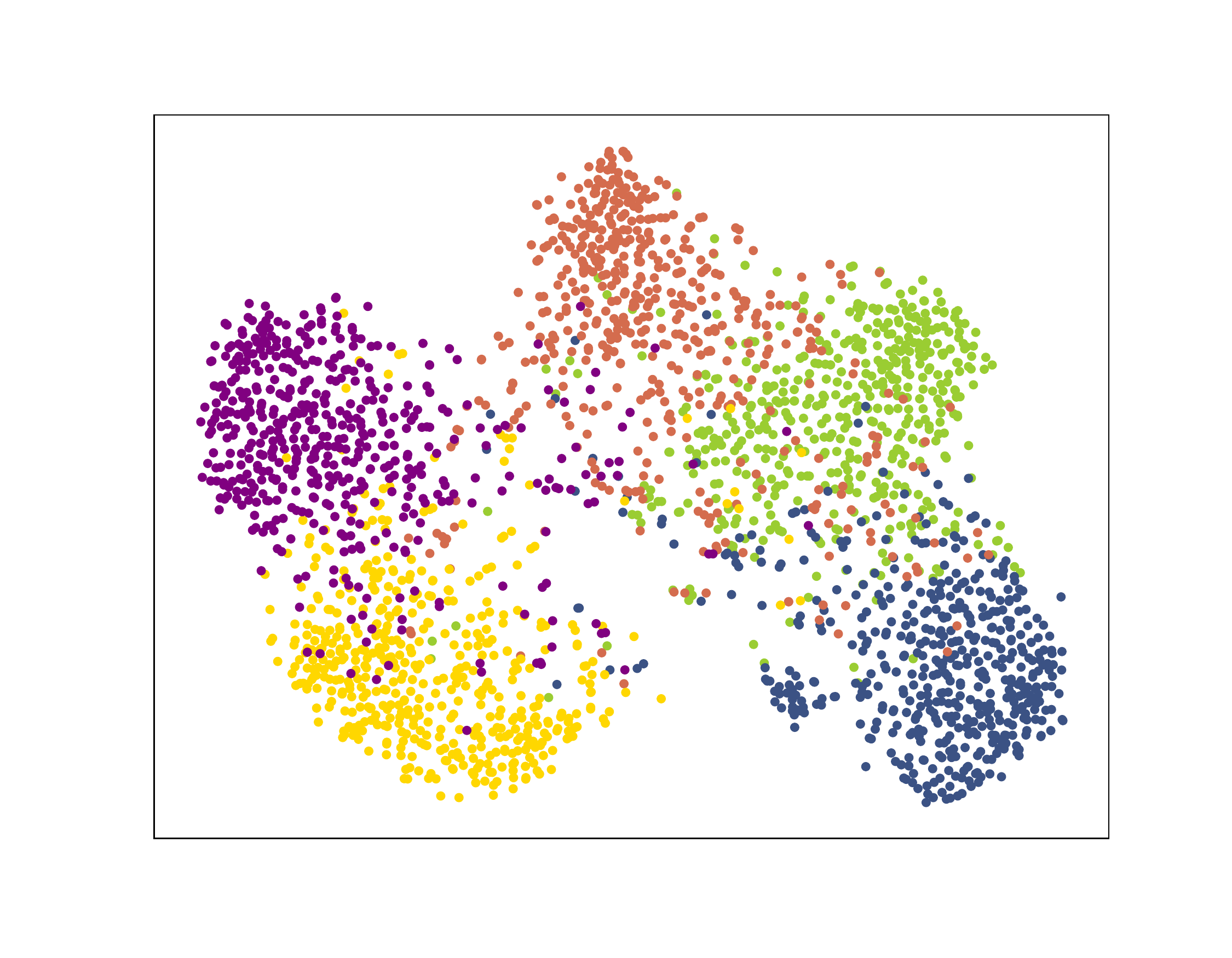}%
    \end{minipage}
    \caption{t-SNE visualization~\cite{van2008visualizing} of representations for simulated CIFAR-10 images.
    }
\label{abla:repre}
\end{figure*}

Recall that RCAL handles noisy labels on long-tailed data based on representation calibration. Here, we visualize achieved representations to demonstrate the effectiveness of the proposed method. To avoid dense figures, we visualize the representations of data points belonging to tail classes. The results are presented in Figure~\ref{abla:repre}.  As
can be seen, RCAL can obtain more robust representations and therefore better classification performance.

\vspace{0.2cm}

\section{Details of Baselines}
Below, we introduce the exploited baselines.
\begin{itemize}
    \item ERM. Using the standard Cross Entropy (CE) loss, the network is simply trained on noisy and imbalanced datasets.
\end{itemize}
\myPara{Methods for long-tailed distributions:}
\begin{itemize}[itemsep=2pt,topsep=0pt,parsep=0pt]
    \item LDAM~\cite{cao2019learning}. This work designs a label-distribution-aware loss function at the class level, which finds the best trade-off between per-class margins.
    \item LDAM-DRW~\cite{cao2019learning}. To overcome the issues brought by re-weighting or re-sampling, a deferred re-balancing training schedule is applied with the LDAM loss. It first trains deep networks with all examples using the LDAM loss with the same weights and then deploys a re-weighted LDAM loss to weigh up the minority classes' losses.
    \item CRT~\cite{kang2019decoupling}. This work claims that data imbalance will not affect the acquisition of high-quality representations, while the strong long-tailed recognition can be achieved by adjusting only the classifier. The learning process is decoupled into representation learning and classifier learning. Representation learning can be performed by different sampling strategies. CRT is to re-train the classifier with class-balanced sampling.
    \item NCM~\cite{kang2019decoupling}. Compared to CRT, NCM is to learn the classifier by computing the mean features representation for each class and then performing the nearest neighbor search. 
    \item MiSLAS~\cite{zhong2021improving}.
\end{itemize}
\myPara{Methods for learning with noisy labels:}
\begin{itemize} [itemsep=2pt,topsep=0pt,parsep=0pt]
    \item Co-teaching~\cite{han2018co}. Two networks are exploited to handle noisy labels simultaneously, which select underlying clean examples for peer networks. 
    \item CDR~\cite{xia2020robust}. Inspired by the lottery ticket hypothesis, this work divides all parameters into critical and non-critical ones. Different types of parameters would perform different update rules to enhance the memorization effect and improve the robustness.
    \item Sel-CL+~\cite{li2022selective}. To learn robust representations, this paper extends supervised contrastive learning by selecting confident pairs. With the learned representations, they further fine tune the classifier.
\end{itemize}
\myPara{Methods for tackling noisy labels
on long-tailed data:}
\begin{itemize}[itemsep=2pt,topsep=0pt,parsep=0pt]
    \item HAR-DRW~\cite{cao2020heteroskedastic}. This work proposes a regularization technique to handle noisy labels and class-imbalanced data in a unified way. Different regularization strength is assigned to each data point, where data point with high uncertainty and low density will be assigned larger regularization strength. 
    \item RoLT~\cite{wei2021robust}. To distinguish mislabeled examples from rare examples, this paper designs a class-dependent noise detector by computing the distance to prototypes. This paper also employs semi-supervised methods to improve the robustness further.
    
    \item RoLT-DRW~\cite{wei2021robust}. Compared to RoLT, a deferred re-weighting technique~\cite{cao2019learning} is leveraged to favor tail classes. 
\end{itemize}

\end{document}